\pdfoutput=1

\documentclass[11pt]{article}

\usepackage[preprint]{acl}

\usepackage{times}
\usepackage{latexsym}
\usepackage{graphicx}
\usepackage{booktabs}    
\usepackage{caption}     
\usepackage{float}       
\usepackage{amsmath}     
\usepackage{multirow}    
\usepackage{caption}
\usepackage{amssymb}
\newcommand{\tmark}{$\triangle$}  
\usepackage{pifont}  

\usepackage{subcaption}
\usepackage{enumitem}
\definecolor{lightgray}{gray}{0.95}
\definecolor{midgray}{gray}{0.90}

\usepackage{booktabs}
\usepackage{multirow}
\usepackage{pifont} 

\newcommand{\cmark}{\ding{51}} 
\newcommand{\xmark}{\ding{55}} 

\usepackage[T1]{fontenc}

\usepackage[utf8]{inputenc}
\usepackage{cuted}
\usepackage{float}
\usepackage{microtype}
\usepackage[most]{tcolorbox}
\usepackage{adjustbox} 
\usepackage{courier}   
\usepackage{inconsolata}

\usepackage[utf8]{inputenc}
\usepackage{amsmath, amssymb}
\usepackage[linesnumbered,ruled,vlined]{algorithm2e}

\usepackage{graphicx}
\usepackage{authblk}

\title{
Mirror in the Model: Ad Banner Image Generation \\
via Reflective Multi-LLM and Multi-modal Agents
}

\author{
  \textbf{Zhao Wang}\textsuperscript{1*}, 
  \textbf{Bowen Chen}\textsuperscript{1,2*}, 
  \textbf{Yotaro Shimose}\textsuperscript{1*}, \\
  \textbf{Sota Moriyama}\textsuperscript{1,3}, 
  \textbf{Heng Wang}\textsuperscript{1}, 
  \textbf{Shingo Takamatsu}\textsuperscript{1} \\
  {\small \textsuperscript{1}Sony Group Corporation, Japan} \\
  {\small \textsuperscript{2}The University of Tokyo, Japan} \\
  {\small \textsuperscript{3}The Graduate University for Advanced Studies, Japan}
}

\begin{document}
\maketitle

\begin{strip}
  \centering
  \includegraphics[width=\linewidth, trim=10 145 10 140, clip]{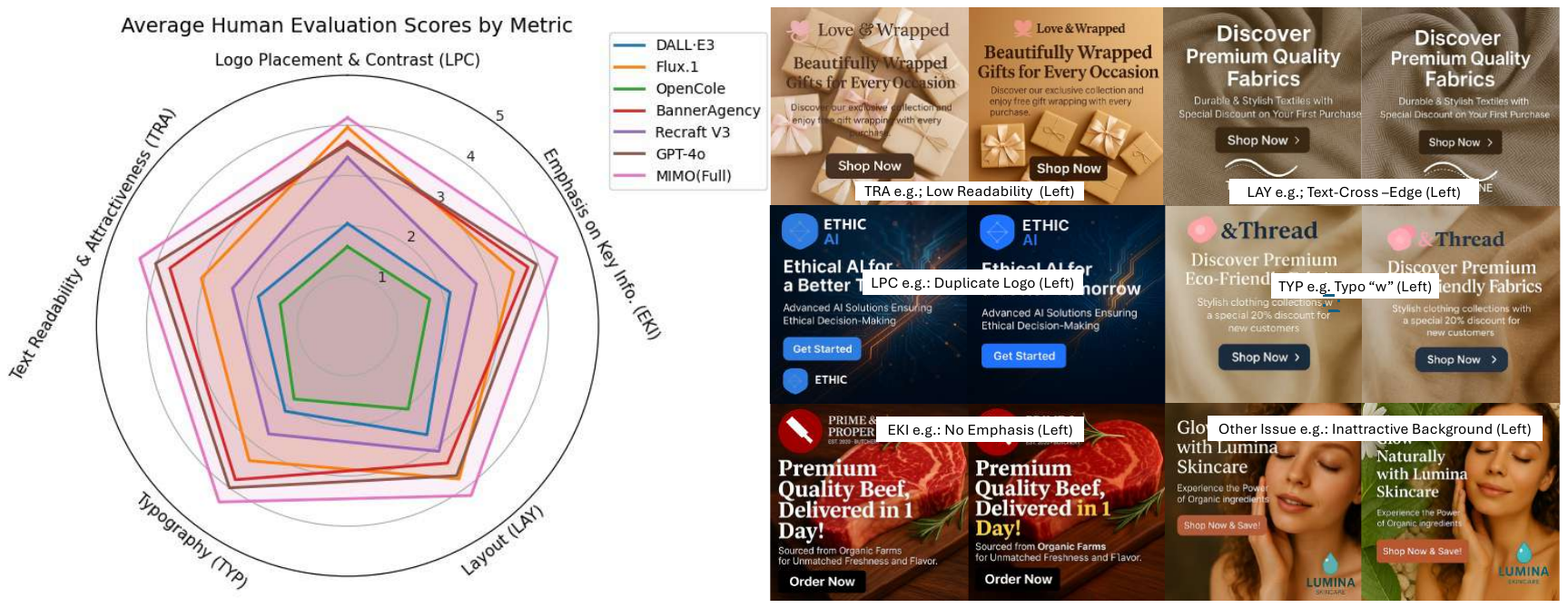}
  \captionof{figure}{
    \textbf{Left}: Radar chart showing human evaluations across baselines and our \textbf{MIMO} for six major design criteria (evaluated on 20 samples), all of which are critical factors in the commercialization of banner ads.
    \textbf{Right}: Six examples of common errors alongside corresponding MIMO refinement. Even the strongest baseline, GPT-4o, produces minor yet critical flaws that are unacceptable in commercial applications.
  }
  \label{fig:font_page_teaser}
\end{strip}

    
\begingroup
\renewcommand\thefootnote{}\footnotetext{\noindent \textsuperscript{*} indicates equal contribution. Bowen Chen and Sota Moriyama completed this work during their internship at Sony. Corresponding author: Zhao.Wang@sony.com}
\endgroup

\begin{abstract}
Recent generative models such as GPT‑4o have shown strong capabilities in producing high-quality images with accurate text rendering. However, commercial design tasks like advertising banners demand more than visual fidelity—they require structured layouts, precise typography, consistent branding and etc. In this paper, we introduce \textbf{MIMO (Mirror In‑the‑Model)}, an agentic refinement framework for automatic ad banner generation. MIMO combines a hierarchical multi-modal agent system (MIMO‑Core) with a coordination loop (MIMO‑Loop) that explores multiple stylistic directions and iteratively improves design quality. Requiring only a simple natural language based prompt and logo image as input, MIMO automatically detects and corrects multiple types of errors during generation. Experiments show that MIMO significantly outperforms existing diffusion and LLM-based baselines in real-world banner design scenarios.
\end{abstract}

\section{Introduction}
\label{sec:intro}
Recent advances in generative modeling have greatly improved image synthesis quality. 
Text-to-image generative models such as DALL·E~\cite{openai2023dalle3} and Flux.1 series~\cite{flux2024} produce high-fidelity visuals. However, these models still face problems such as instruction adherence and text rendering. Hybrid approaches such as OpenCOLE~\cite{opencole2024} and BannerAgency~\cite{wang2025banneragency} have proposed using text-to-image generative models for the background, while employing LLMs to control the layout and attributes of foreground elements such as typography to overcome the weakness of text-to-image generative models. The most recently multimodal LLMs such as GPT‑4o~\cite{openai2024gpt4card} achieve superior semantic–visual alignment~\cite{zhang2025strict}, outperforming existing text-to-image models like Recraft V3~\cite{recraft2024v3} and Flux.1 Pro~\cite{flux2024} in character-level accuracy, instruction adherence, and text rendering. 

However, commercial tasks such as poster or advertising-banner design demand more than visual fidelity—they require structured layout, precise typography, and consistent brand styling. To assess whether existing models meet these standards, we conducted a pretest by manually annotating 20 samples from each baseline (Flux.1 Kontext Max~\cite{flux2025kontextmax}, Recraft V3~\cite{recraft2024v3}, OpenCOLE~\cite{opencole2024}, and BannerAgency~\cite{wang2025banneragency}). We tracked frequent design errors, including \textit{Logo Placement \& Contrast(LPC)}, \textit{Emphasis on Key Info.(EKI)}, \textit{Text Readability \& Attractiveness(TRA)}, \textit{Typography(TYP)} and \textit{Layout(LAY)}. As illustrated in Fig.~\ref{fig:font_page_teaser}, these baselines exhibit diverse and recurring failure modes, highlighting their limitations for real-world advertising applications. Some visualized failures are shown in Fig.~\ref{fig:visual_comparision_baseline} in Appendix~\ref{app:algo}.

Recent breakthroughs in agentic, self‑refinement frameworks—such as SELF‑REFINE, which enables LLMs to provide feedback and iteratively refine outputs~\cite{madaan2023selfrefine}, and TalkHier, a hierarchical multi‑agent feedback system for structured task correction~\cite{wang2025talkhier}—highlight the potential of feedback-driven, step‑by‑step generation refinement.
In the vision domain, Idea2Img~\cite{yang2024idea2imgiterativeselfrefinementgpt4vision} leverages GPT-4V to iteratively edit and refine images through prompt generation, selection, and feedback loops. 
COLE~\cite{jia2024colehierarchicalgenerationframework} fine-tunes LLaVA~\cite{liu2023visualinstructiontuning} to provide visual feedback for refining typographic layout and attributes.
However, neither method adopts an agentic workflow that actively refines the intermediate visual output in a structured and goal-driven manner. 
\medskip
\noindent\fbox{%
  \parbox{\columnwidth}{%
    \textbf{Motivating Question:}  
    \emph{Can we design an agentic module within banner generation models that automatically detects and corrects the visual errors shown at Figure~\ref{fig:font_page_teaser} while also enhancing overall design quality?}  
  }%
}

\medskip
This leads to our proposed \textbf{MIMO (Mirror In‑the‑Model)} architecture. MIMO automatically and continuously inspects intermediate banner drafts, identifies flawed regions—such as misaligned text, missing logos, or inconsistent style—and iteratively refines them, effectively mirroring expert human design review until the output meets professional advertising standards. Specifically, MIMO introduces four key innovations:

\begin{enumerate}[leftmargin=*, nosep, topsep=0pt]
    \item \textbf{MIMO-Core}: a hierarchical, multimodal multi-agent system based on LLMs, which dynamically and autonomously refines visual elements through iterative design–evaluate–revise loops.
    
    \item \textbf{MIMO-Loop}: a high-level coordination layer that simultaneously initiates multiple MIMO-Core instances to explore different stylistic directions. It adopts a multi-agent voting protocol to identify and discard poor designs, while enabling information sharing across cores to boost overall design quality.
    
    \item \textbf{Minimal User Burden}: MIMO requires only a simple natural language prompt and a logo image as user input. All refinement, decision-making, and stylistic exploration are handled automatically—greatly lowering the barrier for commercial deployment.

\end{enumerate}

\section{Methodology}

We consider the task of generating high-quality ad banner images for a product \( R \), given a user-issued prompt \( u \) and a logo image \( \ell \in \mathbb{R}^{H \times W \times 3} \). The prompt \( u \) serves as the primary input for ad generation and may range from a minimal description (e.g., a few keywords like ``summer sale'' or ``premium skincare'') to more detailed advertiser intent or constraints (e.g., ``include 30\% OFF,'' ``show product clearly,'' ``make it premium''), depending on user preference. The logo \( \ell \) must be faithfully embedded into the final design output.

\begin{figure*}[t]
    \centering
    \includegraphics[width=\linewidth, trim=40 240 160 65, clip]{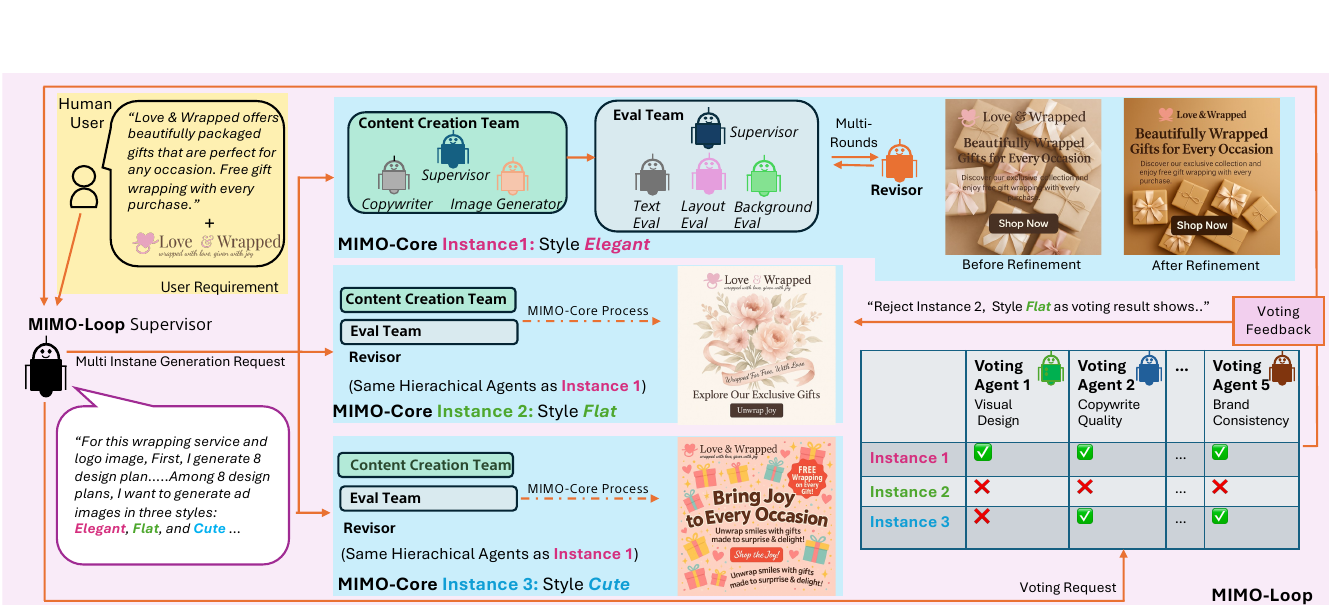}
    \caption{System overview of our proposed \textbf{MIMO} framework. \textbf{MIMO-Core} handles a dynamic refinement process including content generation, evaluation, and revision via a hierarchy of agents. \textbf{MIMO-Loop} wraps around the core to introduce style prompting and multi-agent voting, and iterative refinement for diversity-aware generation.}
    \label{fig:system_overview}
\end{figure*}

MIMO aims to generate a set of \( n \) stylistically diverse ad banner images \( \{x_1, x_2, \dots, x_n\} \), where each \( x_i \in \mathbb{R}^{H' \times W' \times 3} \) satisfies both the semantic intent encoded in \( u \) and the visual branding constraint enforced by \( \ell \). MIMO requires only a simple natural language prompt \( u \) and a logo image \( \ell \) as user input. All refinement, decision-making, and stylistic exploration are handled automatically—greatly lowering the barrier for commercial deployment. Figure~\ref{fig:system_overview} illustrates the overall structure of the MIMO architecture.

\subsection{MIMO-Core: Dynamic Multi-modal, Multi-agents Generate-Refine Pipeline}

The \textbf{MIMO-Core} subsystem (light blue area in Figure~\ref{fig:system_overview}) is a hierarchical multi-agent pipeline that generates a single high-quality banner image through iterative creation, evaluation, and revision. It mimics the structure of human design teams, where agent roles are divided across content creation, quality evaluation, and targeted refinement.

\vspace{0.5em}
\noindent\textbf{Agent Structure.}
We define the agent set:
\[
\mathcal{A}_{\text{core}} = \left\{ \mathcal{A}_{\text{sup}}, \mathcal{A}_{\text{create}}, \mathcal{A}_{\text{eval}}, \mathcal{A}_{\text{rev}} \right\}
\]
with:
\begin{itemize}[leftmargin=*, itemsep=0pt, topsep=0pt]
    \item \( \mathcal{A}_{\text{sup}} \): MIMO-Core Supervisor.
    \item \( \mathcal{A}_{\text{create}} = \{\mathcal{A}_{\text{cs}}, \mathcal{A}_{\text{cw}}, \mathcal{A}_{\text{ig}}, \mathcal{A}_{\text{layout}}\} \): Content Creation Team with a supervisor and three agents: copywriter, image generator, and layout planner.
    \item \( \mathcal{A}_{\text{eval}} = \{\mathcal{A}_{\text{es}}, \mathcal{A}_{\text{text}}, \mathcal{A}_{\text{bg}}, \mathcal{A}_{\text{lay}}\} \): Evaluation Team with a supervisor and agents for text, background, and layout critique.
    \item \( \mathcal{A}_{\text{rev}} \): Graphic Revisor Agent.
\end{itemize}

\vspace{0.5em}
\noindent\textbf{Interaction Protocol.}
At each step \( t \), the system maintains a banner draft \( x^{(t)} \in \mathbb{R}^{H' \times W' \times 3} \) and shared context memory \( \mathcal{M}^{(t)} \). The supervisor selects an active subteam:
\[
\mathcal{A}^{(t)}_{\text{active}} = \text{Route}\left(x^{(t)}, \mathcal{M}^{(t)}\right)
\]
which performs one of three actions:
\begin{itemize}[leftmargin=*, itemsep=0pt,topsep=0pt]
    \item \textbf{Create:} Generate an initial draft via \( \mathcal{A}_{\text{create}} \).
    \item \textbf{Evaluate:} Return textual feedback via \( \mathcal{A}_{\text{eval}} \).
    \item \textbf{Revise:} Refine banner draft to incorporate feedback via \( \mathcal{A}_{\text{rev}} \).
\end{itemize}

Evaluation agents produce feedback messages:
\[
f^{(t)} = \left\{(a_i, \texttt{comment}_i)\right\}_{i=1}^{k}, \quad a_i \in \mathcal{A}_{\text{eval}}
\]
which are aggregated by the Evaluation Team Supervisor and routed to \( \mathcal{A}_{\text{sup}} \). The supervisor then determines whether to re-enter the creation phase, initiate revision, or perform another evaluation.

\noindent\textbf{Multimodal Reasoning.}
All agents operate over multimodal input:
\[
\left(u, \ell, x^{(t)}, \mathcal{M}^{(t)}\right)
\]
where \( u \) is the prompt, \( \ell \) is the logo image, and \( x^{(t)} \) is the current visual draft. Agents reason not only over text but also over rendered designs, enabling grounded, perceptual feedback (e.g., ``CTA overlaps background'' or ``font lacks emphasis'').

This iterative loop continues:
\[
x^{(t)} \rightarrow \text{Eval} \rightarrow \text{Revise} \rightarrow x^{(t+1)}
\]
until the banner is deemed complete by the supervisor \( \mathcal{A}_{\text{sup}} \).

\subsection{MIMO-Loop: Outer-Loop for Style Diversity and Robust Selection}

The \textbf{MIMO-Loop} subsystem builds on MIMO-Core by introducing a diversity-enhancing outer loop that explores multiple stylistic directions and iteratively refines them through agent-based evaluation and elimination. It increases the robustness and expressiveness of the final ad banner while preserving brand fidelity and user intent.

\vspace{0.5em}
\noindent\textbf{Style Prompting.}
Given a logo–prompt pair \( (\ell, u) \), a language model generates \( k \) distinct stylistic prompts:
\[
\mathcal{S} = \{s_1, s_2, \dots, s_k\}
\]
Each style \( s_i \) reflects attributes such as product category, color palette, and branding tone. A style selection agent selects a subset of \( n \) styles:
\[
\mathcal{S}^* = \{s_{i_1}, s_{i_2}, \dots, s_{i_n}\} \subseteq \mathcal{S}
\]
based on the compatibility between style descriptions and the visual–semantic content of \( (\ell, u) \).

\vspace{0.5em}
\noindent\textbf{Parallel MIMO-Core Execution.}
Each selected style prompt \( s_j \in \mathcal{S}^* \) is fed into an independent MIMO-Core instance, resulting in a set of \( n \) refined banner outputs:
\begin{align*}
\mathcal{X}^{(0)} &= \{x_1^{(0)}, x_2^{(0)}, \dots, x_n^{(0)}\} \\
x_j^{(0)} &\gets \texttt{MIMO-Core}(u, \ell, s_j) \quad \text{for } s_j \in \mathcal{S}^*
\end{align*}

\vspace{0.5em}
\noindent\textbf{Multi-Agent Judging.}
Let \( \mathcal{J} = \{\text{Judge}_1, \dots, \text{Judge}_5\} \) denote five specialized \texttt{JudgeAgents}, each evaluating every \( x_j^{(t)} \in \mathcal{X}^{(t)} \) with respect to a distinct criterion (e.g., visual design, copywriting quality, brand consistency, user experience, or technical fidelity). Each judge issues a tuple:
\begin{align*}
r_{j}^{(t)} &= (\texttt{vote}, \texttt{reason}) \\
\texttt{vote} &\in \{\texttt{RECOMMENDED}, \texttt{REJECTED}\}, \quad \texttt{reason} \in \mathcal{T}
\end{align*}

\vspace{0.5em}
\noindent\textbf{Feedback Aggregation and Elimination.}  
Let \( \#\texttt{REJECTED}(x_j^{(t)}) \) denote the number of judges who vote \texttt{REJECTED} for candidate \( x_j^{(t)} \). We eliminate the candidate with the highest rejection count, i.e., \( \mathcal{X}^{(t+1)} = \mathcal{X}^{(t)} \setminus \{x_{j^*}^{(t)}\} \), where \( j^* = \arg\max_j \#\texttt{REJECTED}(x_j^{(t)}) \). 

\vspace{0.5em}
\noindent\textbf{Iterative Refinement Loop.}
This process continues iteratively:
\[
\mathcal{X}^{(t)} \xrightarrow{\text{Judge}} \text{Feedback} 
\xrightarrow{\text{Refine}} \mathcal{X}^{(t+1)}
\]
until \( |\mathcal{X}^{(t)}| = 1 \), at which point the final banner \( x^* \in \mathcal{X}^{(t)} \) is selected as the winner.

The complete procedural steps of both the \texttt{MIMO-Core} and \texttt{MIMO-Loop} pipelines—including agent initialization, hierarchical routing, style diversification, and iterative refinement—are detailed in Algorithm~\ref{alg:mimo-core} and Algorithm~\ref{alg:mimo-loop} in Appendix~\ref{app:algo}.

\noindent
\textbf{Logo and Product Image Authenticity.} To ensure the authenticity of logos and product images in the generated banners, we use external inpainting tools~\cite{recraft2024v3} for manual replacement. Please refer to Appendix~\ref{logo} for further details.

\section{Experimental Result}

\textbf{Datasets.}
We use the \textit{BannerRequest400} dataset from the BannerAgency ~\cite{wang2025banneragency}, which contains 100 real-world advertising prompts with product metadata, logo images, and human-designed banners. The dataset spans categories such as electronics, cosmetics, and home goods. Each sample includes a product name, marketing description, campaign constraints, and logo—supporting evaluation of both text-to-image generation and layout-aware design.

\noindent
\textbf{Baselines.}
We compare MIMO framework with five baselines covering latest text-to-image models, multi-modal LLM and hybrid approaches.
\begin{itemize}[leftmargin=0.3em, itemsep=0pt, topsep=0pt, parsep=0pt]
\item \textbf{DALL·E 3}~\cite{openai2023dalle3}: A proprietary text-to-image generative model that produces images from natural language prompts.

\item \textbf{Flux.1 Kontext Max}~\cite{flux2025kontextmax}: The latest flow-matching model in the Flux.1 family. It offers enhanced prompt adherence and controllability.

\item \textbf{Recraft V3}~\cite{recraft2024v3}: A commercial text-to-image model specialized in text rendering.

\item \textbf{OpenCole}~\cite{opencole2024}: An open-source hybrid system that combines the background generation of diffusion models with the reasoning ability of language models.

\item \textbf{BannerAgency}~\cite{wang2025banneragency}: A multi-modal LLM agent framework that generates advertisement banners using a static agent workflow.

\item \textbf{GPT‑4o}~\cite{openai2024gpt4card}: OpenAI’s latest multimodal model capable of generating and editing images from natural language prompts with strong semantic–visual alignment.
\end{itemize}

\begin{table*}[h]
\centering
\scriptsize
\begin{tabular}{l@{\hskip 12pt}c@{\hskip 10pt}c@{\hskip 10pt}c@{\hskip 10pt}c@{\hskip 10pt}c@{\hskip 10pt}c@{\hskip 10pt}c@{\hskip 10pt}c@{\hskip 10pt}c@{\hskip 10pt}c@{\hskip 10pt}c}
\toprule
\textbf{Method} & \textbf{TAA} & \textbf{LPS} & \textbf{CTAE} & \textbf{CPYQ} & \textbf{BIS} & \textbf{AQS} & \textbf{LPC} & \textbf{EKI} & \textbf{LAY} & \textbf{TYP} & \textbf{TRA} \\
\midrule
DALL·E3        & 3.17$^{0.19}$ & 1.78$^{0.21}$ & 4.18$^{0.20}$ & 2.48$^{0.18}$ & 3.06$^{0.22}$ & 3.10$^{0.20}$ & 3.10$^{0.35}$ & 2.95$^{0.40}$ & 3.25$^{0.38}$ & 3.00$^{0.42}$ & 3.05$^{0.39}$ \\
Flux.1        & 3.36$^{0.28}$ & 3.67$^{0.26}$ & 4.28$^{0.25}$ & 2.64$^{0.27}$ & 3.09$^{0.25}$ & 3.12$^{0.26}$ & 4.54$^{0.70}$ & 3.96$^{1.72}$ & 4.27$^{1.13}$ & 3.99$^{1.78}$ & 4.00$^{1.67}$ \\
OpenCole        & 2.10$^{0.20}$ & 1.20$^{0.22}$ & 2.30$^{0.21}$ & 2.40$^{0.20}$ & 2.25$^{0.19}$ & 2.50$^{0.18}$ & 2.85$^{0.41}$ & 2.70$^{0.45}$ & 2.90$^{0.44}$ & 2.60$^{0.50}$ & 2.75$^{0.48}$ \\
BannerAgency    & 3.30$^{0.45}$ & 3.10$^{0.42}$ & 3.40$^{0.40}$ & 3.25$^{0.43}$ & 3.20$^{0.41}$ & 3.35$^{0.44}$ & 3.96$^{1.74}$ & 3.64$^{1.40}$ & 3.80$^{1.21}$ & 3.83$^{1.17}$ & 3.80$^{1.15}$ \\
Recraft V3    & 3.85$^{0.29}$ & 3.75$^{0.27}$ & 3.90$^{0.26}$ & 3.72$^{0.28}$ & 3.80$^{0.30}$ & 3.88$^{0.27}$ & 4.46$^{0.93}$ & 3.86$^{1.95}$ & 4.07$^{1.79}$ & 3.88$^{2.01}$ & 3.90$^{1.90}$ \\
GPT-4o          & 4.15$^{0.30}$ & 4.10$^{0.28}$ & 4.20$^{0.27}$ & 4.10$^{0.29}$ & 4.15$^{0.30}$ & 4.18$^{0.28}$ & 4.97$^{0.04}$ & 4.90$^{0.14}$ & 4.88$^{0.19}$ & 4.81$^{0.46}$ & 4.84$^{0.26}$ \\
\midrule
\textbf{MIMO-Core(-)} & 4.48$^{0.21}$ & 4.41$^{0.20}$ & 4.29$^{0.19}$ & 4.37$^{0.22}$ & 4.50$^{0.18}$ & 4.53$^{0.20}$ & 4.95$^{0.06}$ & 4.93$^{0.08}$ & 4.92$^{0.08}$ & 4.89$^{0.09}$ & 4.90$^{0.08}$ \\
\textbf{MIMO-Core}  & 4.62$^{0.18}$ & 4.60$^{0.19}$ & 4.45$^{0.20}$ & 4.58$^{0.21}$ & 4.66$^{0.22}$ & 4.71$^{0.19}$ & 4.97$^{0.05}$ & 4.96$^{0.06}$ & 4.96$^{0.05}$ & 4.95$^{0.06}$ & 4.95$^{0.05}$ \\
\textbf{MIMO(Full)} & \textbf{4.83}$^{0.17}$ & \textbf{4.81}$^{0.18}$ & \textbf{4.65}$^{0.19}$ & \textbf{4.70}$^{0.18}$ & \textbf{4.78}$^{0.19}$ & \textbf{4.88}$^{0.18}$ & \textbf{4.98}$^{0.03}$ & \textbf{4.97}$^{0.04}$ & \textbf{4.97}$^{0.04}$ & \textbf{4.96}$^{0.05}$ & \textbf{4.96}$^{0.04}$ \\
\bottomrule
\end{tabular}
\caption{Comparison of average metric scores across methods (higher is better). \textbf{MIMO-Core(-)} = MIMO without refinement; \textbf{MIMO-Core} = MIMO with refinement; \textbf{MIMO(Full)} = MIMO with refinement and loop.}
\label{tab:main-results}
\end{table*}

\noindent
\textbf{Implementation Details.}  
 For the commercial models (e.g., DALL·E 3, Recraft V3, Flux.1 Kontext Max), we use their public API to generate outputs under default or recommended settings. Prompt design for each baseline model is detailed in Appendix \ref{basepromptdesign}. 
We use GPT‑4o as the backbone LLM for MIMO, with the decoding temperature fixed at 0 across all experiments. MIMO is implemented using the LangGraph\cite{langgraph2024} . All the prompts we used in MIMO are detailed in Appendix~\ref{app:eval_prompt}, \ref{app:single_agent_prompt} and \ref{basepromptdesign}. 

\noindent
\textbf{Tool Usage.} MIMO only employs two tools:

\begin{itemize}[leftmargin=1em,itemsep=0pt,topsep=0pt,parsep=0pt]
  \item \textbf{MultiModal I/O Tool:} Interfaces with the GPT-4o vision model for both image generation and visual input. It enables agents to perceive and analyze images directly, which is critical for layout and design tasks.
  \item \textbf{Output Tool:} A simple tool used to save internal thoughts or logs from agents. It does not return any output, and is mainly for memory tracking during agent communication.
\end{itemize}

\noindent
\textbf{Evaluation Metrics.}  
We evaluate image quality using the following six criteria proposed by BannerAgency\cite{wang2025banneragency} for fair comparison including: \textit{TAA (Target Audience Alignment)}, \textit{LPS (Logo Placement Score)}, \textit{CTAE (Call-to-Action Effectiveness)}, \textit{CPYQ (Copy Quality)}, \textit{BIS(Brand Integration Score)} and \textit{AQS (Aesthetic Quality Score)}.

To check if MIMO can handle the problems mentioned in Sec.~\ref{sec:intro}, we designed the following metrics: 

\begin{description}[
  leftmargin=0em,
  labelsep=0.5em,
  itemsep=0pt,
  parsep=0pt,
  topsep=0pt,
  labelwidth=5em,
  align=left,
  font=\normalfont
]

  \item[\textit{Logo Placement \& Contrast (LPC):}]
  Checks if the logo is easy to see, clearly placed, and stands out from the background. It does not evaluate the authenticity of the logo, as most baseline methods do not support this functionality.

  \item[\textit{Emphasis on Key Information (EKI):}]
  Looks at whether important messages like discounts or key features are highlighted clearly.

  \item[\textit{Layout (LAY):}]
  Evaluates the arrangement of elements, including spacing and alignment, for a clean and balanced design.

  \item[\textit{Typography (TYP):}]
  Checks if the font style, size, and consistency make the text easy to read and visually appealing.

  \item[\textit{Text Readability \& Attractiveness (TRA):}]
  Measures how easy the text is to read and whether it looks good within the design.

\end{description}

\subsection{Comparison with Baselines}

Table~\ref{tab:main-results} compares the average metric scores of various methods across 13 evaluation criteria. Overall, our proposed MIMO variants outperform all baselines. Among them, MIMO(Full) achieves the highest scores across all metrics, demonstrating the effectiveness of both refinement and loop-based optimization. The MIMO-Core version, which includes refinement but excludes the loop mechanism, still performs strongly and consistently surpasses MIMO-Core(-) (the variant without refinement), highlighting the critical role of iterative revision. Notably, while GPT-4o performs competitively in aesthetic and layout-related metrics, it falls short in structured design elements such as Logo Placement (LPC) and Emphasis on Key Information (EKI), where MIMO excels. Other baseline models perform poorly across most dimensions, particularly in structured or brand-sensitive attributes. These results underscore the importance of integrating hierarchical refinement and adaptive feedback loops for generating commercially viable ad creatives. A detailed visualized comparison with baselines is shown in Fig~\ref{fig:visual_comparision_baseline} in Appendix~\ref{app:algo}. Visualization of MIMO-Core Refinement and MIMO-Loop Stylishment are shown in Fig~\ref{fig:mimo_core_refine} and ~\ref{fig:mimo_loop_styles} in Appendix~\ref{sec:visualization}.

\subsection{Human Evaluation}
Vision-language models often struggle to fully understand image content, as noted in~\cite{rahmanzadehgervi2025visionlanguagemodelsblind}. BannerAgency~\cite{wang2025banneragency} showed a strong correlation between human evaluations and GPT-4o-based evaluations using their proposed metrics.
To verify the reliability of GPT-4o-based evaluation on our new metrics, we collected scores from 12 human evaluators on 20\% of the prompts. Table~\ref{tab:human-results} shows the average human scores, which align well with machine-based evaluations, achieving a Spearman’s rank correlation of \textbf{0.85}. MIMO outperforms all baselines, confirming the validity of our evaluation method.

While GPT-based evaluations generally rank models correctly, human evaluators reveal performance gaps more clearly. For instance, weaker models like OpenCole and DALL·E3 receive much lower scores from humans compared to stronger models such as Flux.1 or BannerAgency. In contrast, GPT-4o scores tend to be closer across models. This suggests that human evaluation offers more precise and detailed insights, especially for capturing qualitative differences in generation quality.


\begin{table}
\centering
\scriptsize
\begin{tabular}{l@{\hskip 20pt}c@{\hskip 20pt}c@{\hskip 20pt}c@{\hskip 20pt}c@{\hskip 20pt}c}
\toprule
\textbf{Method} & \textbf{LPC} & \textbf{EKI} & \textbf{LAY} & \textbf{TYP} & \textbf{TRA} \\
\midrule
DALL·E3          & 2.04 & 2.15 & 2.69 & 2.11 & 1.87 \\
Flux.1 & 3.96 & 3.48 & 3.78 & 3.33 & 3.06 \\
OpenCole         & 1.58 & 1.72 & 2.06     & 1.81     & 1.41 \\
BannerAgency    & 3.67 & 3.78 & 3.39     & 3.80 & 3.72 \\
Recraft V3          & 3.36 & 2.70 & 3.10 & 2.67 & 2.41 \\
GPT-4o           & 3.62 & 3.96 & 3.70 & 4.00 & 4.02 \\

\midrule
MIMO(Full)             & \textbf{4.15} & \textbf{4.39} & \textbf{4.19} & \textbf{4.35} & \textbf{4.35} \\
\bottomrule
\end{tabular}
\caption{Average metric scores from human evaluators}
\label{tab:human-results}
\end{table}

\subsection{Ablation Study: Single- vs. Multi-Agent}

To assess the importance of decomposing the ad generation process into multiple agents, we conducted a controlled ablation study comparing a single-agent setup against MIMO-Core framework. The single-agent setting processes all requirements via a single, comprehensive prompt (including all detailed prompts in multi-agent based MIMO-Core setting to ensure fair comparison), while the multi-agent system distributes the generation and evaluation tasks across specialized roles. Each configuration was evaluated over three independent runs. Table~\ref{tab:ablation_single_vs_multi_transposed} shows the averaged metric scores and standard deviations across the three runs. This result highlights the limitation of the single-agent setup: simply packing all requirements into one prompt yields weaker performance. In contrast, the multi-agent MIMO framework assigns tasks to specialized agents in stages—MIMO-Core handles layout, copywriting, and evaluation separately, while MIMO-Loop further refines style—leading to more accurate and higher-quality outputs.

\begin{table}[b]
\centering
\scriptsize
\begin{tabular}{l@{\hskip 6pt}c@{\hskip 6pt}c@{\hskip 6pt}c@{\hskip 6pt}c@{\hskip 6pt}c@{\hskip 6pt}c@{\hskip 6pt}}
\toprule
\textbf{Setting} & \textbf{TAA} & \textbf{LPS} & \textbf{CTAE} & \textbf{CPYQ} & \textbf{BIS} & \textbf{AQS}  \\
\midrule
Single & 4.01$^{0.09}$ & 4.05$^{0.07}$ & 3.89$^{0.10}$ & 3.85$^{0.11}$ & 4.29$^{0.07}$ & 4.01$^{0.09}$  \\
\textbf{Multi} & \textbf{4.37}$^{0.18}$ & \textbf{4.59}$^{0.06}$ & \textbf{4.77}$^{0.17}$ & \textbf{4.66}$^{0.04}$ & \textbf{4.76}$^{0.13}$ & \textbf{4.79}$^{0.05}$  \\
\bottomrule
\end{tabular}
\caption{Ablation Study: GPT-4o Scores for Single-Agent vs. Multi-Agent (Mean$^{\text{Std}}$ across 3 runs)}
\label{tab:ablation_single_vs_multi_transposed}
\end{table}

\subsection{Ablation Study: Number of Evaluation Agents in the MIMO-Loop}
In MIMO-Loop, we deploy five evaluation agents to vote on the quality of banners generated from different styles. We examine how performance is affected by the number of evaluation agents, the number of styles, and the importance of each evaluation criterion. 

\begin{figure}[t]
    \centering
    \includegraphics[width=\columnwidth]{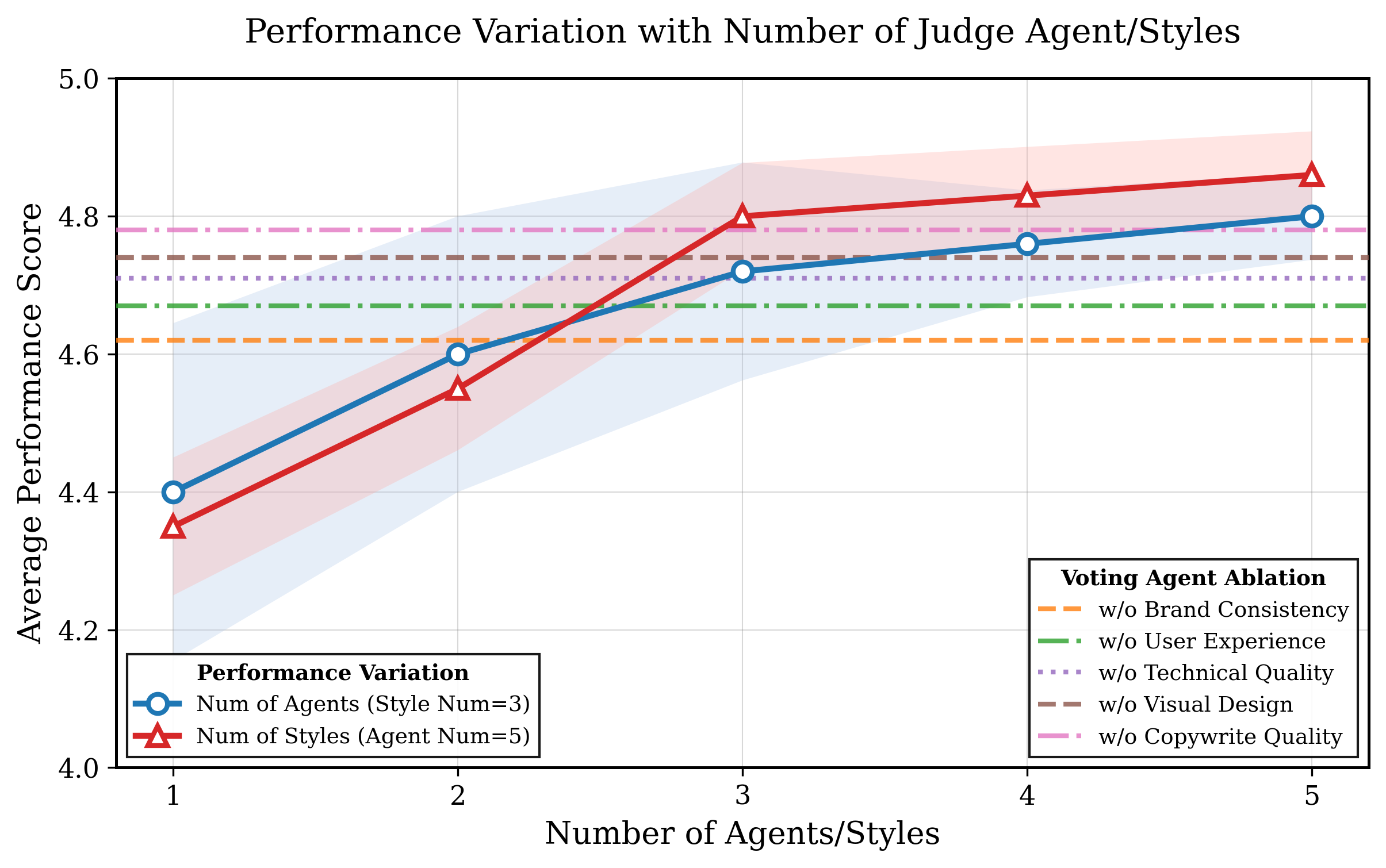}
    \caption{We measure average performance by varying the number of agents or styles (solid lines), while keeping the other fixed. We also run an ablation study by removing one voting agent at a time (dotted lines). }
    \label{fig:mimo_loop_ablation}
\end{figure}

As shown in Figure~\ref{fig:mimo_loop_ablation}, performance improves as the number of evaluation agents increases. Using only one agent results in lower and more variable scores, suggesting that a single agent cannot provide a comprehensive assessment. Adding more agents not only improves performance but also reduces variance, leading to more consistent results. We also observe that increasing the number of styles boosts performance, with a larger gain from 1 to 3 styles than from 3 to 5. Finally, in the voting-agent ablation study, we find that removing the Brand Consistency agent causes the most significant drop in performance, underscoring the importance of preserving logo integrity. In contrast, removing the Copywriting Quality agent has minimal impact.

\section{Related Work}

\paragraph{Image Generation \& Ad Design.}  
Text-to-image models such as DALL·E 3~\cite{openai2023dalle3}, Flux.1 Pro ~\cite{flux2024} and Recraft V3~\cite{recraft2024v3} achieve high visual fidelity but often struggle with structured layout, typography accuracy, and brand-style coherence—especially in banner or poster designs. Multimodal LLM-driven frameworks, including OpenCOLE~\cite{opencole2024}, AutoPoster~\cite{lin2023autoposter}, and BannerAgency~\cite{wang2025banneragency}, improve text rendering and blueprint creation but still exhibit recurring flaws in alignment, emphasis, and brand consistency, as our pretest and Figure~\ref{fig:font_page_teaser} confirm.

\noindent
\textbf{Agentic Refinement.}
Text-only systems like SELF-REFINE~\cite{madaan2023selfrefine} use an LLM to generate feedback and make iterative revisions without extra training. TalkHier~\cite{wang2025talkhier} builds on this by introducing a structured, multi-agent setup with communication rules, shared context, and unbiased evaluation. Idea2Img~\cite{yang2024idea2imgiterativeselfrefinementgpt4vision} leverages GPT-4V to iteratively edit and refine images through prompt generation, selection, and feedback loops. COLE~\cite{jia2024colehierarchicalgenerationframework} fine-tunes LLaVA~\cite{liu2023visualinstructiontuning} to provide visual feedback for refining typographic layout and attributes. 

\section*{Limitations}

While MIMO delivers substantial improvements in banner quality, layout fidelity, and brand alignment via multi-agent coordination, it incurs a higher computational cost. Compared to GPT-4o’s one-shot generation (\$0.21 per image), our full pipeline (MIMO-Core + Loop) may cost up to \$2.87 per banner. A detailed cost breakdown is presented in Appendix~\ref{app:token}. Considering that a visually compelling banner can significantly boost user engagement and conversion, we leave it to practitioners and advertisers to judge whether this cost–quality trade-off is justified in their specific use cases.

Another limitation is the reliance on a manual logo and product image replacement process in this paper. We found that this is the only way to ensure 100\% authenticity of the logo and product image. Automatic replacement methods, such as mask-based inpainting, often fail to preserve the fine details of the original logo and product design. These methods can unintentionally distort elements such as shadows, gradients, or positioning, resulting in visually inconsistent or incorrect integration within the banner. Therefore, to maintain brand integrity, we adopt a manual approach for the replacement.

\bibliography{custom}

\begin{thebibliography}{18}
\providecommand{\natexlab}[1]{#1}

\bibitem[{Flux.1~AI()}]{imagepromptcollection}
Inc. Flux.1~AI.
\newblock Ai image prompt collection.
\newblock \url{https://flux1.ai/prompts}.
\newblock Accessed: 2025-06-20.

\bibitem[{Inoue et~al.(2024)Inoue, Masui, Shimoda, and Yamaguchi}]{opencole2024}
Naoto Inoue, Kento Masui, Wataru Shimoda, and Kota Yamaguchi. 2024.
\newblock \href {https://arxiv.org/abs/2406.08232} {Opencole: Towards reproducible automatic graphic design generation}.
\newblock \emph{Preprint}, arXiv:2406.08232.

\bibitem[{Jia et~al.(2024)Jia, Li, Yuan, Liu, Shen, Chen, Chen, Zheng, Chen, Li, Xie, Zhang, and Guo}]{jia2024colehierarchicalgenerationframework}
Peidong Jia, Chenxuan Li, Yuhui Yuan, Zeyu Liu, Yichao Shen, Bohan Chen, Xingru Chen, Yinglin Zheng, Dong Chen, Ji~Li, Xiaodong Xie, Shanghang Zhang, and Baining Guo. 2024.
\newblock \href {https://arxiv.org/abs/2311.16974} {Cole: A hierarchical generation framework for multi-layered and editable graphic design}.
\newblock \emph{Preprint}, arXiv:2311.16974.

\bibitem[{Labs(2024)}]{flux2024}
Black~Forest Labs. 2024.
\newblock Flux.
\newblock \url{https://github.com/black-forest-labs/flux}.

\bibitem[{Labs(2025)}]{flux2025kontextmax}
Black~Forest Labs. 2025.
\newblock Flux.1-kontext max.
\newblock \url{https://bfl.ai/models/flux-kontext}.
\newblock Accessed: 2025-06-20.

\bibitem[{LangChain(2024)}]{langgraph2024}
LangChain. 2024.
\newblock Langgraph: State machine for llm applications.
\newblock \url{https://github.com/langchain-ai/langgraph}.
\newblock Accessed: 2025-06-26.

\bibitem[{Lin et~al.(2023)Lin, Zhou, Ma, Gao, Fei, Chen, and et~al.}]{lin2023autoposter}
Jinpeng Lin, Min Zhou, Ye~Ma, Yifan Gao, Chenxi Fei, Yangjian Chen, and et~al. 2023.
\newblock Autoposter: A highly automatic and content-aware design system for advertising poster generation.
\newblock In \emph{ACM Multimedia}.

\bibitem[{Liu et~al.(2023)Liu, Li, Wu, and Lee}]{liu2023visualinstructiontuning}
Haotian Liu, Chunyuan Li, Qingyang Wu, and Yong~Jae Lee. 2023.
\newblock \href {https://arxiv.org/abs/2304.08485} {Visual instruction tuning}.
\newblock \emph{Preprint}, arXiv:2304.08485.

\bibitem[{Madaan et~al.(2023)Madaan, Tandon, Gupta, Hallinan, Gao, Wiegreffe, Alon, Dziri, Prabhumoye, Yang, Gupta, Majumder, Hermann, Welleck, Yazdanbakhsh, and Clark}]{madaan2023selfrefine}
Aman Madaan, Niket Tandon, Prakhar Gupta, Skyler Hallinan, Luyu Gao, Sarah Wiegreffe, Uri Alon, Nouha Dziri, Shrimai Prabhumoye, Yiming Yang, Shashank Gupta, Bodhisattwa~Prasad Majumder, Katherine Hermann, Sean Welleck, Amir Yazdanbakhsh, and Peter Clark. 2023.
\newblock Self-refine: Iterative refinement with self-feedback.
\newblock \emph{arXiv preprint arXiv:2303.17651}.

\bibitem[{OpenAI(2023)}]{openai2023dalle3}
OpenAI. 2023.
\newblock {DALL\ensuremath{\cdot}E 3 System Card}.
\newblock \url{https://cdn.openai.com/papers/dall-e-3.pdf}.

\bibitem[{OpenAI(2025)}]{openai2024gpt4card}
OpenAI. 2025.
\newblock Gpt-4o report.
\newblock \url{https://openai.com/ja-JP/index/introducing-4o-image-generation/}.
\newblock Accessed: 2025-06-09.

\bibitem[{Rahmanzadehgervi et~al.(2025)Rahmanzadehgervi, Bolton, Taesiri, and Nguyen}]{rahmanzadehgervi2025visionlanguagemodelsblind}
Pooyan Rahmanzadehgervi, Logan Bolton, Mohammad~Reza Taesiri, and Anh~Totti Nguyen. 2025.
\newblock \href {https://arxiv.org/abs/2407.06581} {Vision language models are blind: Failing to translate detailed visual features into words}.
\newblock \emph{Preprint}, arXiv:2407.06581.

\bibitem[{{Recraft}(2025)}]{recraft_prompt_crafting}
{Recraft}. 2025.
\newblock How to craft prompts for accurate, ai‑generated images.
\newblock \url{https://recraft.ai/blog/}.
\newblock Accessed: 2025-06-20.

\bibitem[{{Recraft Inc.}(2024)}]{recraft2024v3}
{Recraft Inc.} 2024.
\newblock Recraft v3.
\newblock \url{https://www.recraft.ai}.
\newblock Web-based text-to-image model emphasizing long-text rendering, anatomical correctness, and top-rank performance on design-quality leaderboards.

\bibitem[{Wang et~al.(2025{\natexlab{a}})Wang, Shimose, and Takamatsu}]{wang2025banneragency}
Heng Wang, Yotaro Shimose, and Shingo Takamatsu. 2025{\natexlab{a}}.
\newblock Banneragency: Advertising banner design with multimodal llm agents.
\newblock \emph{arXiv}.

\bibitem[{Wang et~al.(2025{\natexlab{b}})Wang, Moriyama, Wang, Gangopadhyay, and Takamatsu}]{wang2025talkhier}
Zhao Wang, Sota Moriyama, Wei‑Yao Wang, Briti Gangopadhyay, and Shingo Takamatsu. 2025{\natexlab{b}}.
\newblock Talk structurally, act hierarchically: A collaborative framework for llm multi-agent systems.
\newblock \emph{arXiv preprint arXiv:2502.11098}.

\bibitem[{Yang et~al.(2024)Yang, Wang, Li, Lin, Lin, Liu, and Wang}]{yang2024idea2imgiterativeselfrefinementgpt4vision}
Zhengyuan Yang, Jianfeng Wang, Linjie Li, Kevin Lin, Chung-Ching Lin, Zicheng Liu, and Lijuan Wang. 2024.
\newblock \href {https://arxiv.org/abs/2310.08541} {Idea2img: Iterative self-refinement with gpt-4v(ision) for automatic image design and generation}.
\newblock \emph{Preprint}, arXiv:2310.08541.

\bibitem[{Zhang et~al.(2025)Zhang, Wang, Tai, Li, Chi, Tian, He, and Wang}]{zhang2025strict}
Tianyu Zhang, Xinyu Wang, Zhenghan Tai, Lu~Li, Jijun Chi, Jingrui Tian, Hailin He, and Suyuchen Wang. 2025.
\newblock Strict: Stress test of rendering images containing text.
\newblock \emph{arXiv preprint arXiv:2505.18985}.
\newblock Evaluates GPT‑4o, Gemini 2.0, Recraft V3, FLUX 1.1 Pro on text‑rendering accuracy and instruction compliance.

\end{thebibliography}

\appendix
\onecolumn

\section{Algorithm of MIMO-Core and MIMO-Loop; Visualized Comparison with Baselines}
\label{app:algo}

Algorithm~\ref{alg:mimo-core} presents the full procedure of the MIMO-Core pipeline, which governs the generation of a high-quality banner image through iterative collaboration among hierarchical multi-agent teams. The algorithm reflects dynamic team selection, internal routing within content creation and evaluation subteams, and multimodal feedback refinement. Each agent acts over shared memory (default setting in Langraph~\cite{langgraph2024} )and rendered visual drafts, and the loop continues until the supervisor determines that the output meets quality standards.

\begin{algorithm}[h]
\caption{MIMO-Core: Dynamic Multimodal Generate-Refine Pipeline}
\label{alg:mimo-core}
\KwIn{Prompt \( u \), Logo image \( \ell \)}
\KwOut{Final banner image \( x^{(T)} \)}

Initialize shared memory \( \mathcal{M}^{(0)} \gets \{u, \ell\} \)\;
Initialize visual draft \( x^{(0)} \gets \emptyset \), step \( t \gets 0 \)\;
Initialize agent hierarchy:\;
\quad MIMO-Core Supervisor \( \mathcal{A}_{\text{sup}} \)\;
\quad Content Creation Team \( \mathcal{A}_{\text{create}} = \{\mathcal{A}_{\text{cs}}, \mathcal{A}_{\text{cw}}, \mathcal{A}_{\text{ig}}, \mathcal{A}_{\text{layout}}\} \)\;
\quad Evaluation Team \( \mathcal{A}_{\text{eval}} = \{\mathcal{A}_{\text{es}}, \mathcal{A}_{\text{text}}, \mathcal{A}_{\text{bg}}, \mathcal{A}_{\text{lay}}\} \)\;
\quad Graphic Revisor Agent \( \mathcal{A}_{\text{rev}} \)\;

\While{\( \mathcal{A}_{\text{sup}} \)\ decides not terminated}{
    Supervisor \( \mathcal{A}_{\text{sup}} \) selects active team:
    \(\mathcal{A}^{(t)}_{\text{team}} \gets \text{Route}(x^{(t)}, \mathcal{M}^{(t)})\);

    \uIf{\( \mathcal{A}^{(t)}_{\text{team}} = \mathcal{A}_{\text{create}} \)}{
        Team supervisor \( \mathcal{A}_{\text{cs}} \) selects agent(s):\;
        \quad \( \mathcal{A}_{\text{create}}^{(t)} \gets \text{Route}_{\text{create}}(\mathcal{M}^{(t)}) \)\;
        Active creation agent(s) generate content:\;
        \quad \( x^{(t+1)} \gets \text{Create}(u, \ell, \mathcal{M}^{(t)}) \)\;
    }
    \uElseIf{\( \mathcal{A}^{(t)}_{\text{team}} = \mathcal{A}_{\text{eval}} \)}{
        Team supervisor \( \mathcal{A}_{\text{es}} \) selects agent(s):\;
        \quad \( \mathcal{A}_{\text{eval}}^{(t)} \gets \text{Route}_{\text{eval}}(x^{(t)}) \)\;
        Active evaluation agents return feedback:\;
        \quad \( f^{(t)} \gets \text{Evaluate}(x^{(t)}, \mathcal{M}^{(t)}) \)\;
        Update shared memory:\;
        \quad \( \mathcal{M}^{(t+1)} \gets \mathcal{M}^{(t)} \cup f^{(t)} \)\;
    }
    \uElseIf{\( \mathcal{A}^{(t)}_{\text{team}} = \mathcal{A}_{\text{rev}} \)}{
        Graphic Revisor applies revisions:\;
        \quad \( x^{(t+1)} \gets \text{Revise}(x^{(t)}, f^{(t)}, \mathcal{M}^{(t)}) \)\;
    }

    \( t \gets t + 1 \)\;
}
\Return \( x^{(T)} \)\;
\end{algorithm}

\clearpage
Algorithm~\ref{alg:mimo-loop} presents the full procedure of the MIMO-Loop subsystem, which builds on MIMO-Core to enhance design diversity and robustness. It begins by generating multiple style prompts based on the initial input \( (u, \ell) \), then selects a subset of stylistically distinct directions. Each style prompt is processed independently by a parallel instance of MIMO-Core. The resulting banners are evaluated by a panel of JudgeAgents, who issue both a vote and a reason. The most-rejected banner is iteratively removed, and feedback is passed to the corresponding MIMO-Core agents for refinement. This elimination-and-refinement loop continues until a single final banner is selected.

\begin{algorithm}[h]
\caption{MIMO-Loop: Style Diversity and Iterative Selection}
\label{alg:mimo-loop}
\KwIn{Prompt \( u \), Logo image \( \ell \), total style pool size \( k \), number of selected styles \( n \)}
\KwOut{Final banner image \( x^* \)}

Generate \( k \) stylistic prompts:
\[
\mathcal{S} \gets \{s_1, s_2, \dots, s_k\}
\]

Select a subset of \( n \) styles:
\[
\mathcal{S}^* \gets \text{SelectStyles}(\mathcal{S}, u, \ell)
\]

Initialize style candidates via parallel MIMO-Core calls:\;
\ForEach{\( s_j \in \mathcal{S}^* \)}{
    \( x_j^{(0)} \gets \texttt{MIMO-Core}(u, \ell, s_j) \)\;
}
\( \mathcal{X}^{(0)} \gets \{x_1^{(0)}, x_2^{(0)}, \dots, x_n^{(0)}\} \), step \( t \gets 0 \)\;

\While{\( |\mathcal{X}^{(t)}| > 1 \)}{
    JudgeAgents evaluate each \( x_j^{(t)} \in \mathcal{X}^{(t)} \) and issue:
    \[
    r_j^{(t)} = (\texttt{vote}, \texttt{reason})
    \]

    Count \texttt{REJECTED} votes for each candidate\;
    Identify index:
    \[
    j^* \gets \arg\max_j \#\texttt{REJECTED}(x_j^{(t)})
    \]

    Eliminate worst candidate:
    \[
    \mathcal{X}^{(t+1)} \gets \mathcal{X}^{(t)} \setminus \{x_{j^*}^{(t)}\}
    \]

    Pass feedback to MIMO-Core instance \( j^* \) for refinement\;
    \( t \gets t + 1 \)\;
}
\Return final banner \( x^* \in \mathcal{X}^{(t)} \)
\end{algorithm}

\label{sec:visualization_baseline}
\begin{figure*}[h]
    \centering
    \includegraphics[width=\linewidth, trim=0 60 0 60, clip]{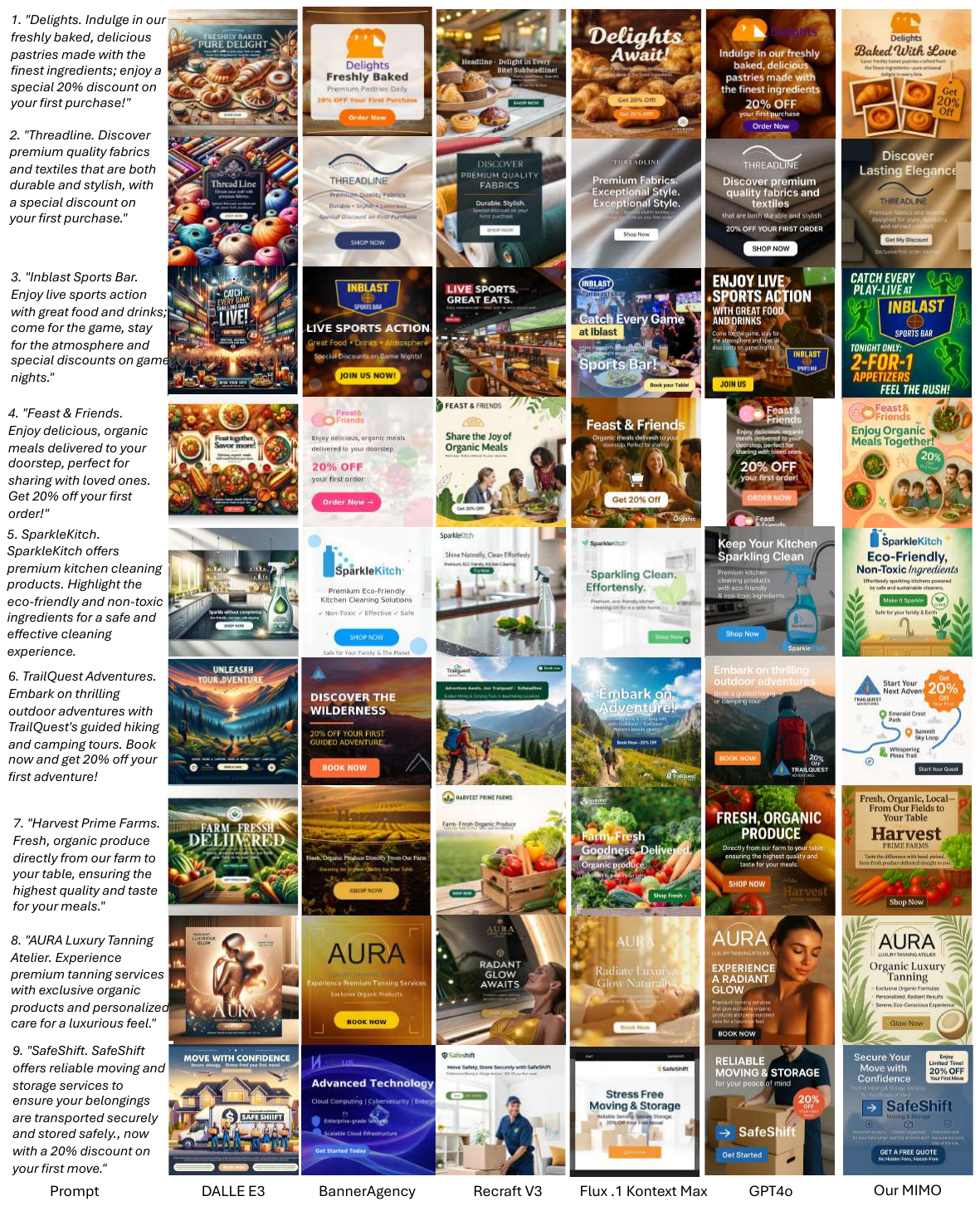}
    \caption{Visual comparison between our method and baseline models. Even the strongest baseline—GPT-4o, the latest image generation model—exhibits notable failures across various prompts: poor layout design (Prompt 6), misleading ad text (Prompt 2, where ``20\% off'' is omitted despite being specified), typographical errors (e.g., ``ecclusive'' in Prompt 8), low logo contrast (Prompts 1 \& 7), incorrect aspect ratio (Prompt 4), and prompt misinterpretation (Prompt 9, where a storage-moving service is mistaken for luggage moving). In contrast, our proposed MIMO framework—combining multi-agent refinement (MIMO-Core) with stylistic enhancement (MIMO-Loop)—consistently produces visually appealing and prompt-faithful banner images.}
    \label{fig:visual_comparision_baseline}
\end{figure*}

\clearpage
\section{Visualization of MIMO-Core Refinement and MIMO-Loop Stylishment }
\label{sec:visualization}

\begin{figure*}[h]
    \centering
    \begin{subfigure}{0.22\textwidth}
        \includegraphics[width=\linewidth]{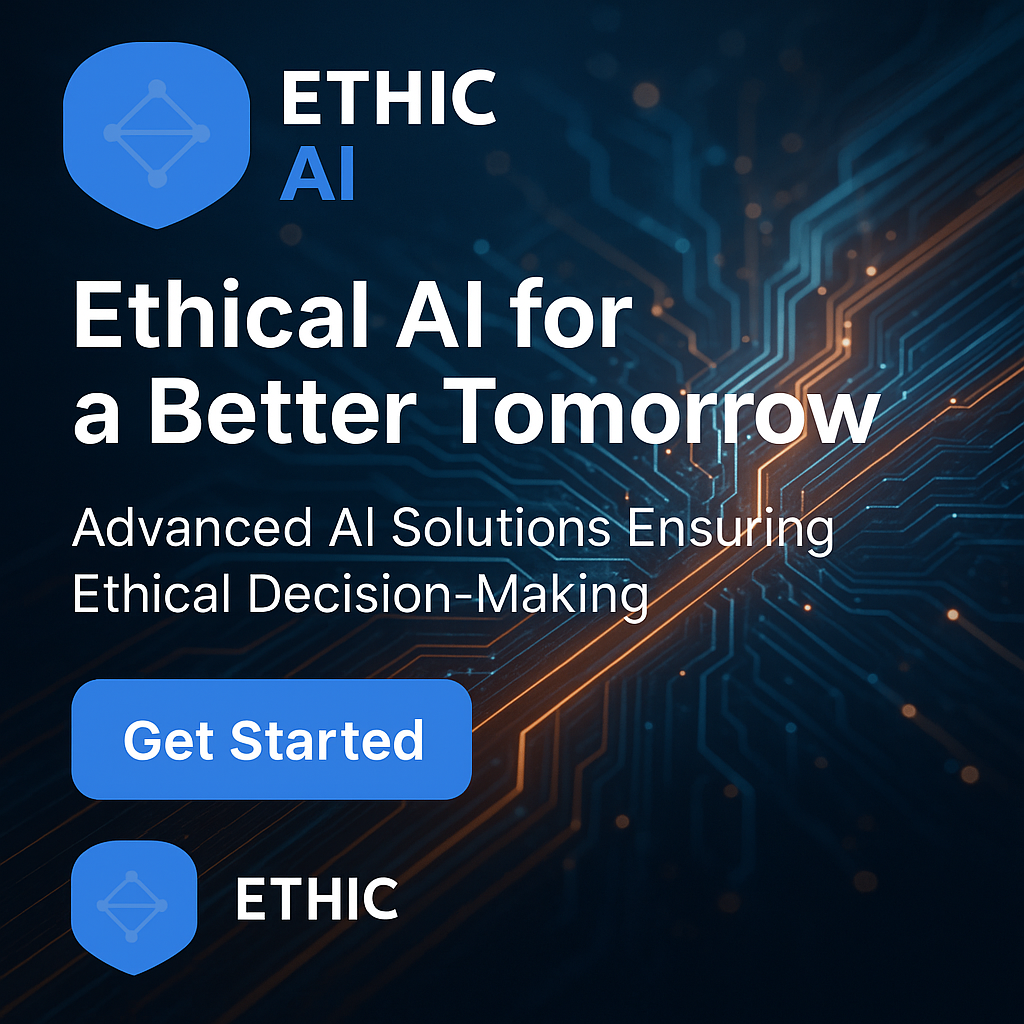}
        \caption*{\textbf{(a1)} Duplicate Logos}
    \end{subfigure}
    \hspace{0.01\textwidth}
    \begin{subfigure}{0.22\textwidth}
        \includegraphics[width=\linewidth]{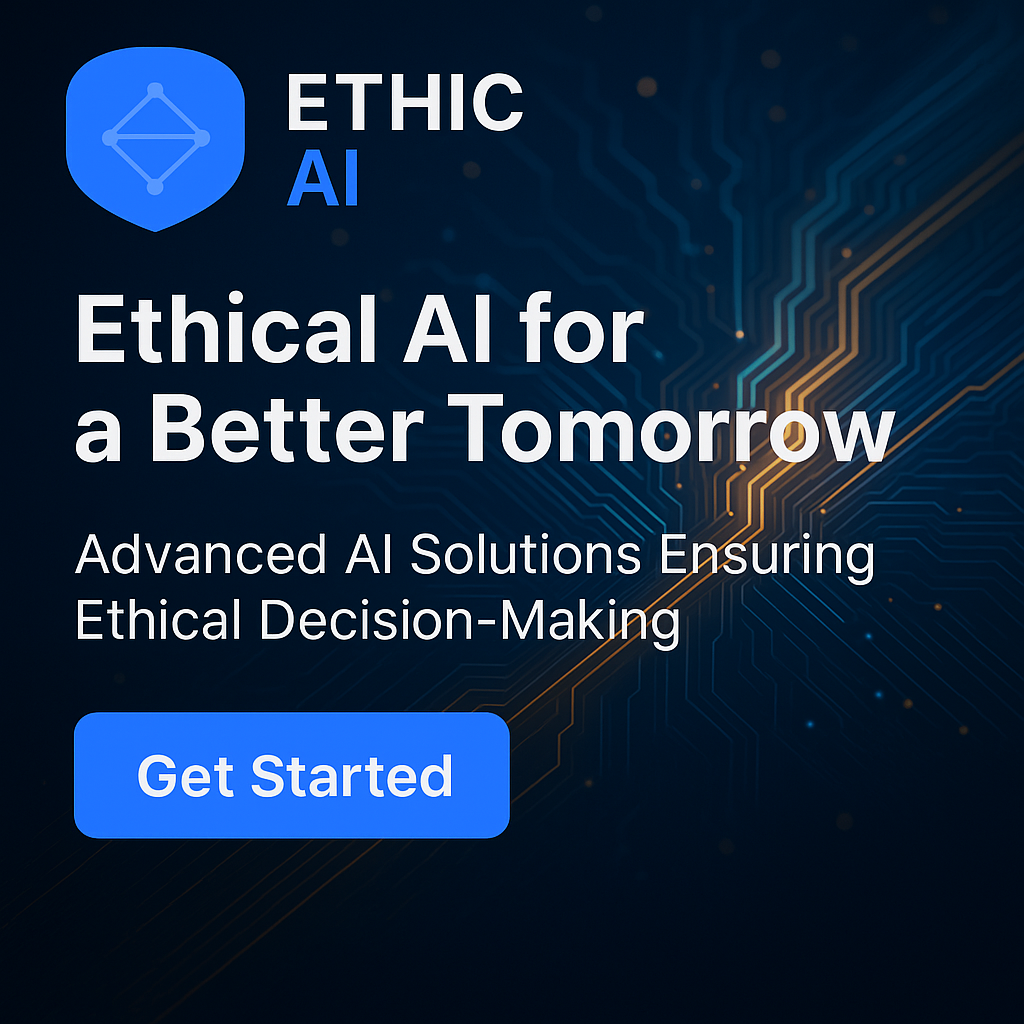}
        \caption*{\textbf{(a2)} After Refinement}
    \end{subfigure}
    \hspace{0.01\textwidth}
    \begin{subfigure}{0.22\textwidth}
        \includegraphics[width=\linewidth]{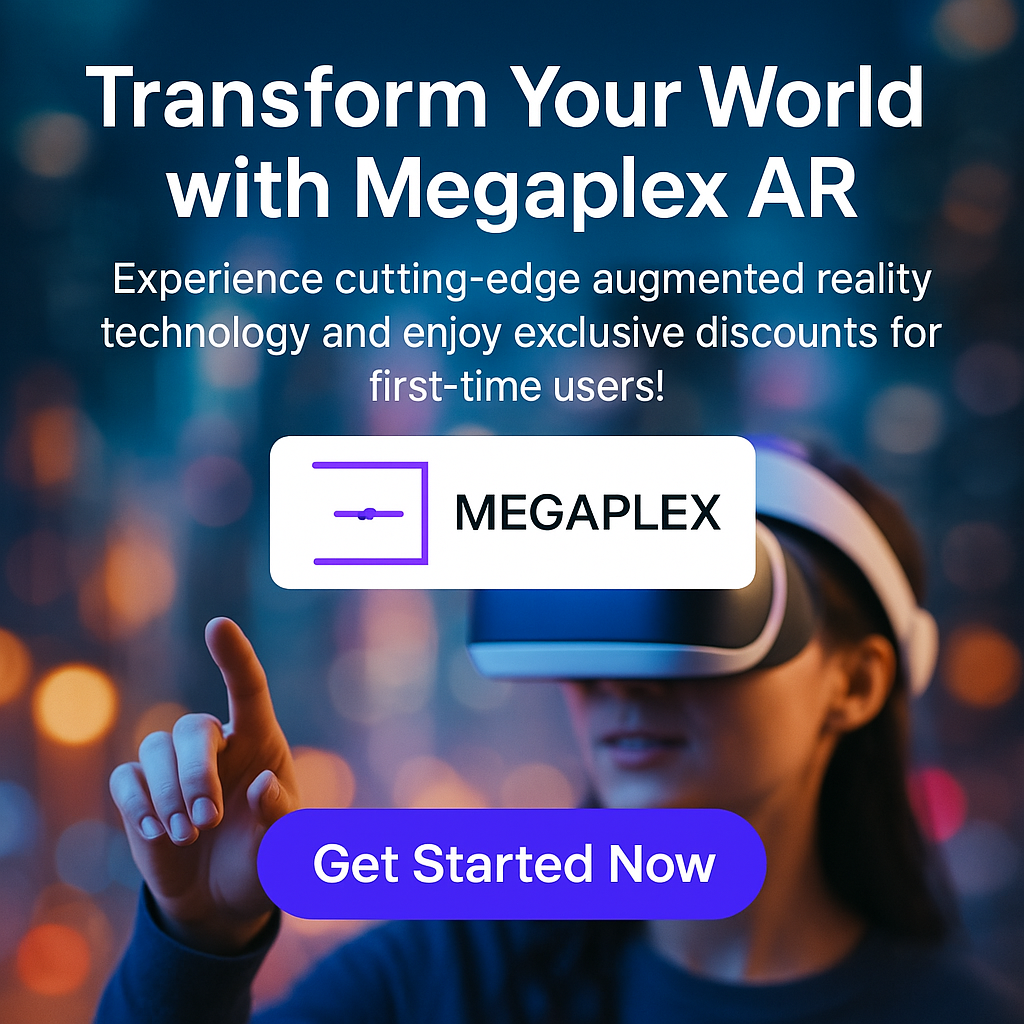}
        \caption*{\textbf{(b1)} Logo Placement}
    \end{subfigure}
    \hspace{0.01\textwidth}
    \begin{subfigure}{0.22\textwidth}
        \includegraphics[width=\linewidth]{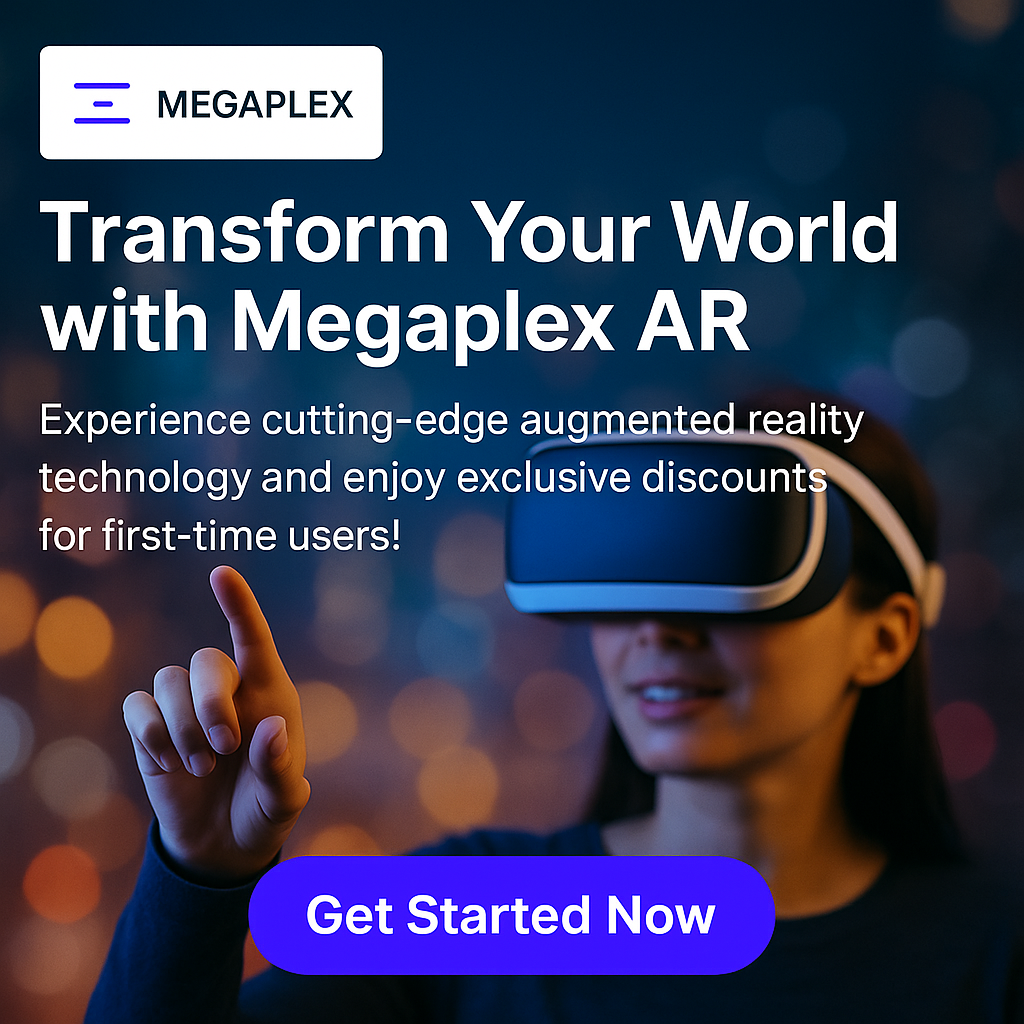}
        \caption*{\textbf{(b2)} After Refinement}
    \end{subfigure}

    \begin{subfigure}{0.22\textwidth}
        \includegraphics[width=\linewidth]{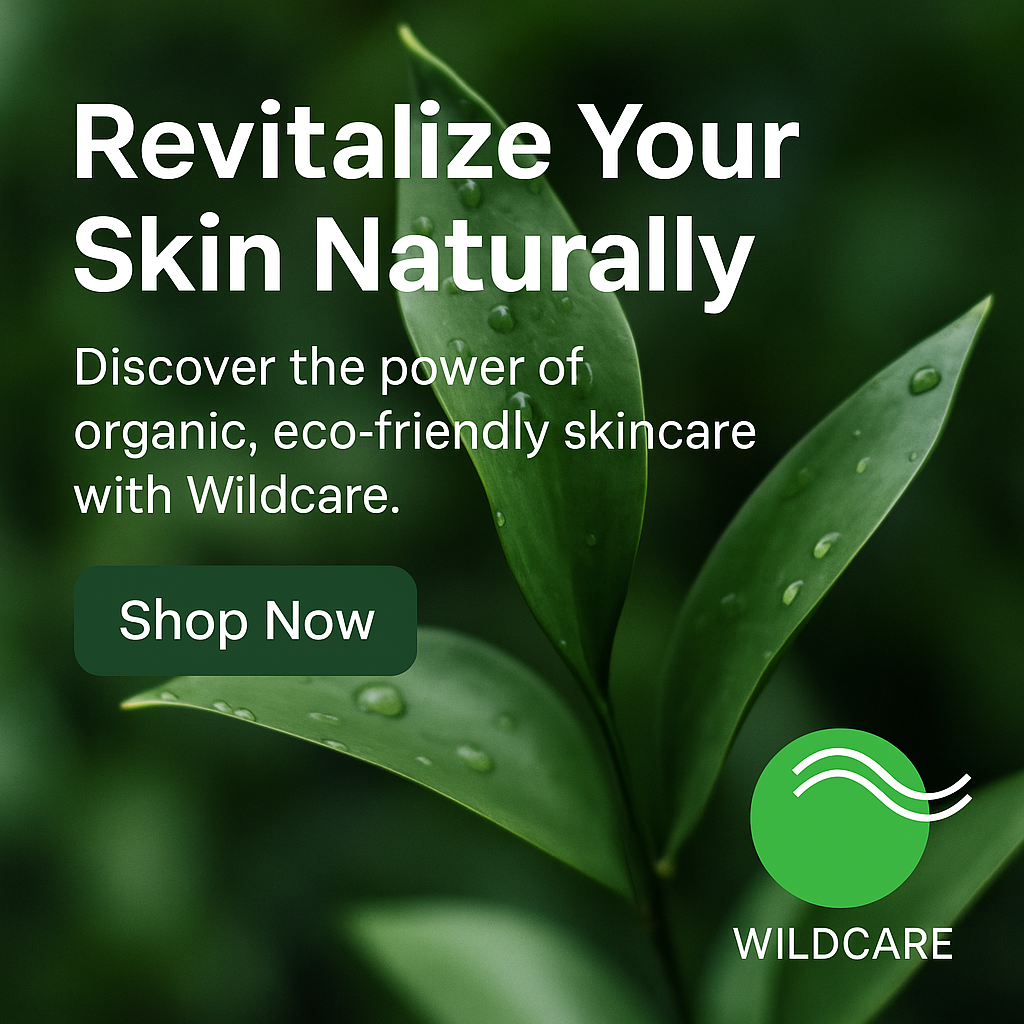}
        \caption*{\textbf{(c1)} Imbalanced Scale}
    \end{subfigure}
    \hspace{0.01\textwidth}
    \begin{subfigure}{0.22\textwidth}
        \includegraphics[width=\linewidth]{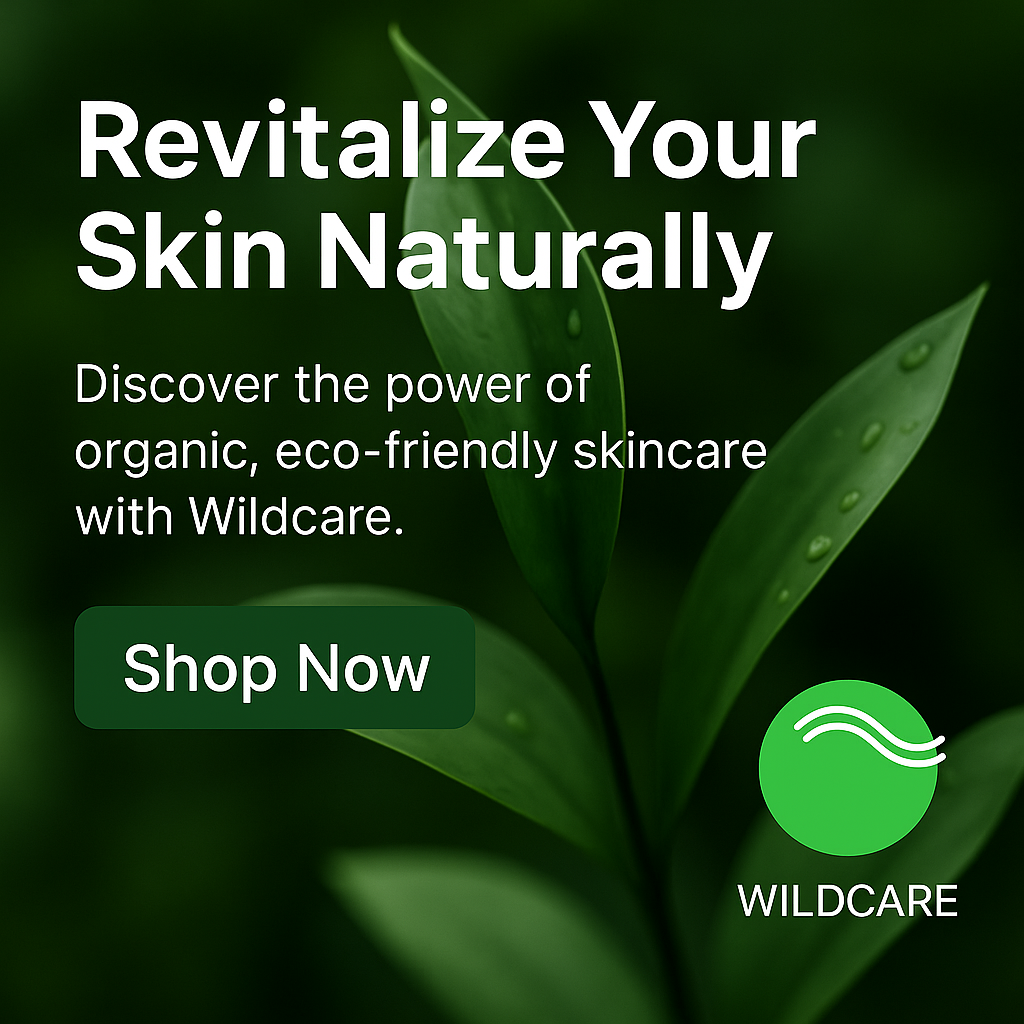}
        \caption*{\textbf{(c2)} After Refinement}
    \end{subfigure}
    \hspace{0.01\textwidth}
    \begin{subfigure}{0.22\textwidth}
        \includegraphics[width=\linewidth]{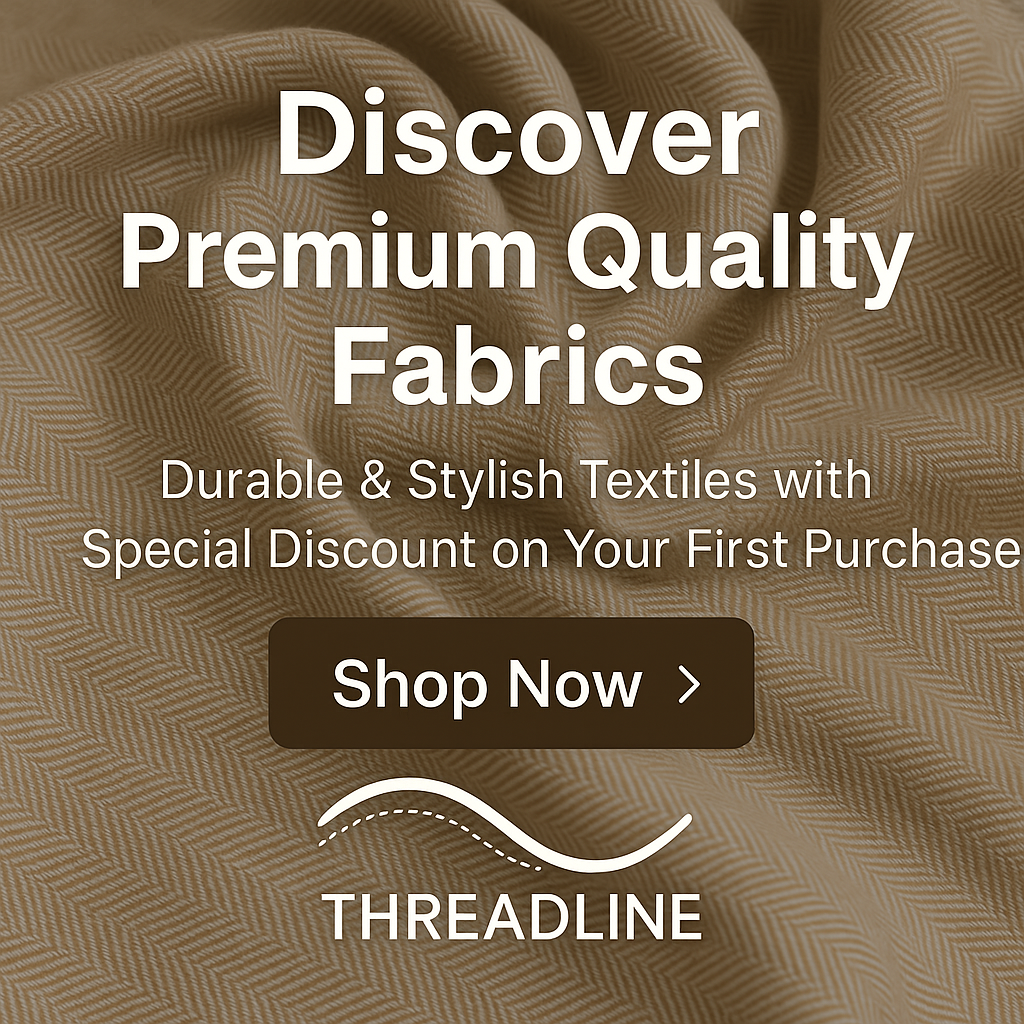}
        \caption*{\textbf{(d1)} Text Crossing Margin}
    \end{subfigure}
    \hspace{0.01\textwidth}
    \begin{subfigure}{0.22\textwidth}
        \includegraphics[width=\linewidth]{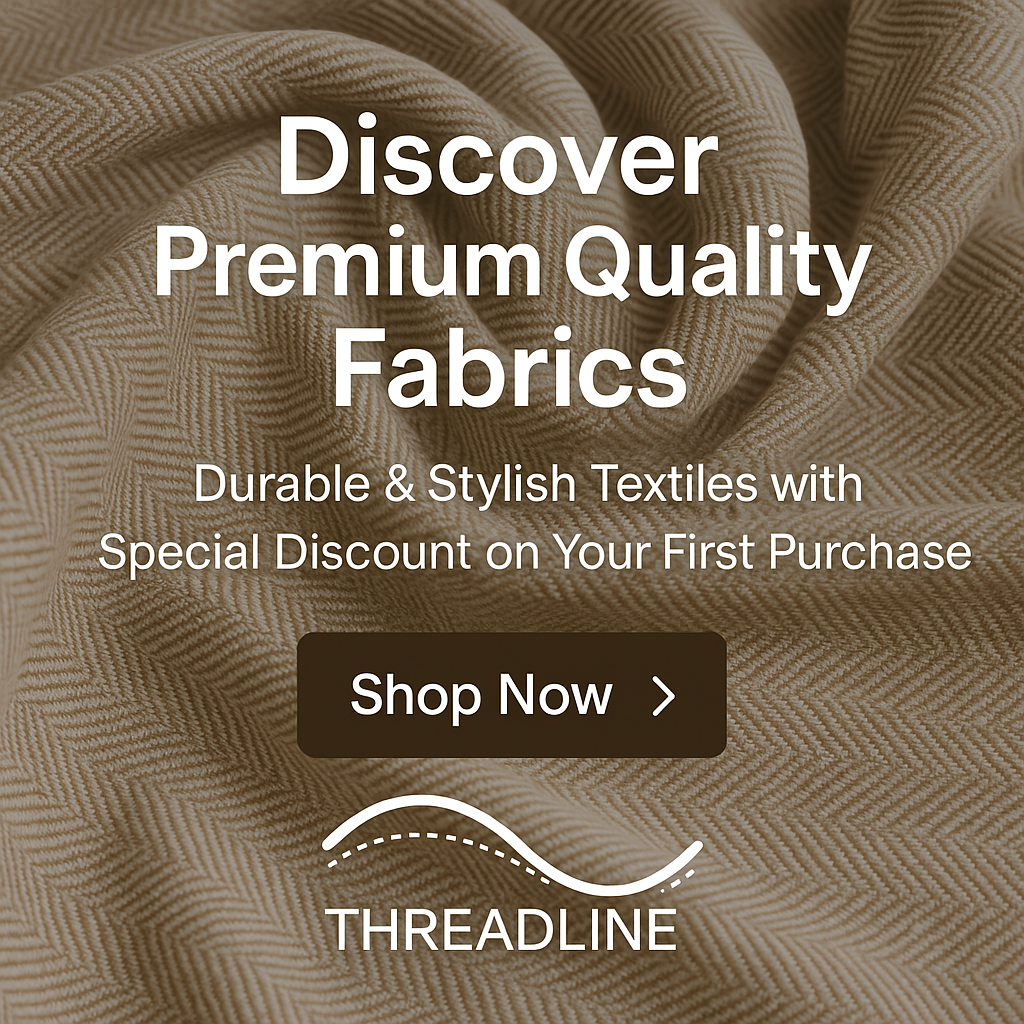}
        \caption*{\textbf{(d2)} After Refinement}
    \end{subfigure}

    \begin{subfigure}{0.22\textwidth}
        \includegraphics[width=\linewidth]{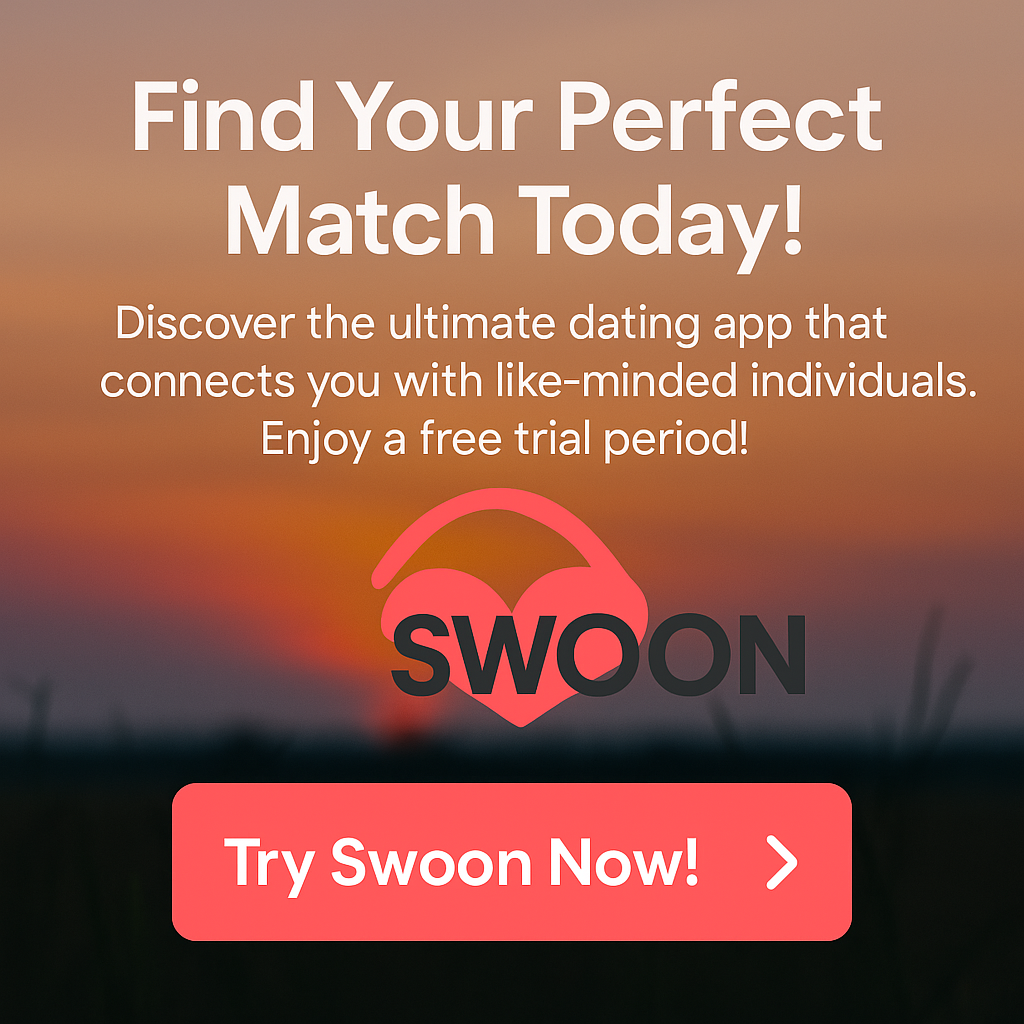}
        \caption*{\textbf{(e1)} Poor Background}
    \end{subfigure}
    \hspace{0.01\textwidth}
    \begin{subfigure}{0.22\textwidth}
        \includegraphics[width=\linewidth]{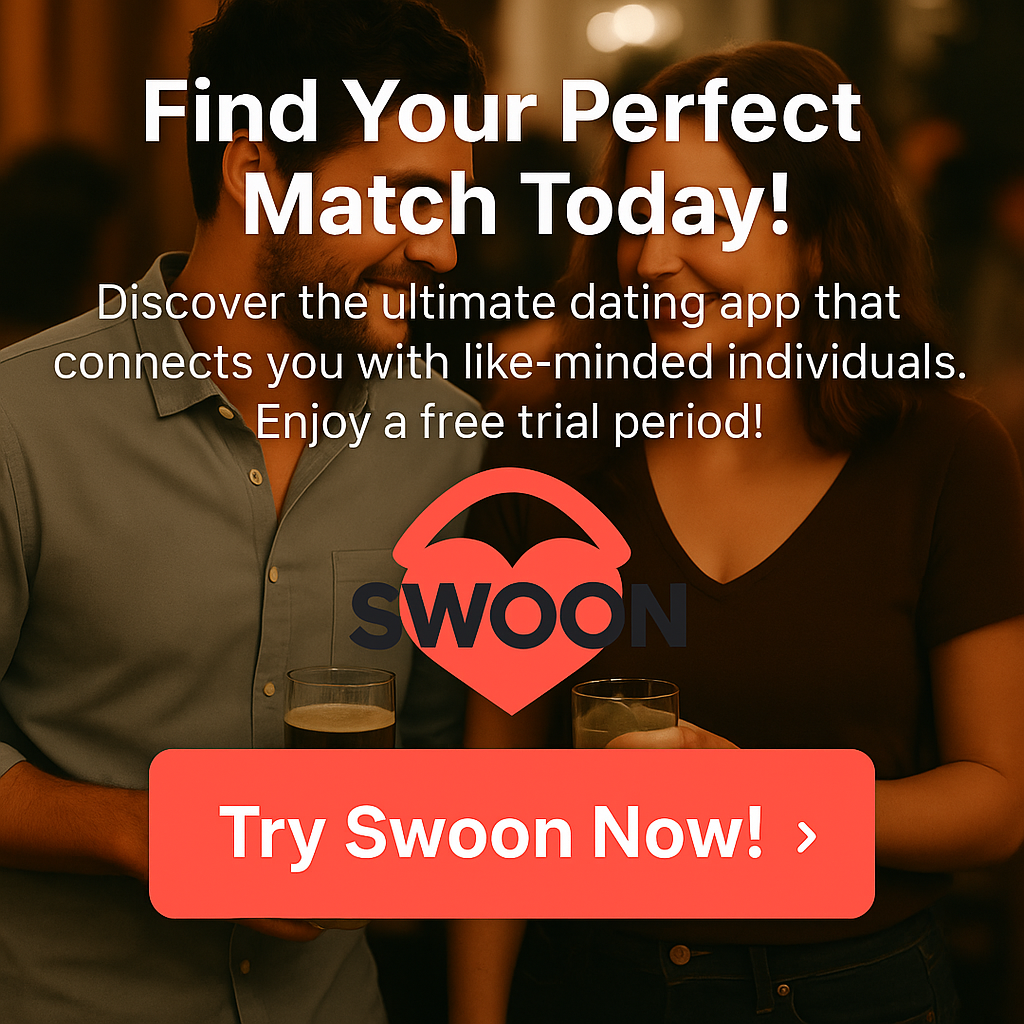}
        \caption*{\textbf{(e2)} After Refinement}
    \end{subfigure}
    \hspace{0.01\textwidth}
    \begin{subfigure}{0.22\textwidth}
        \includegraphics[width=\linewidth]{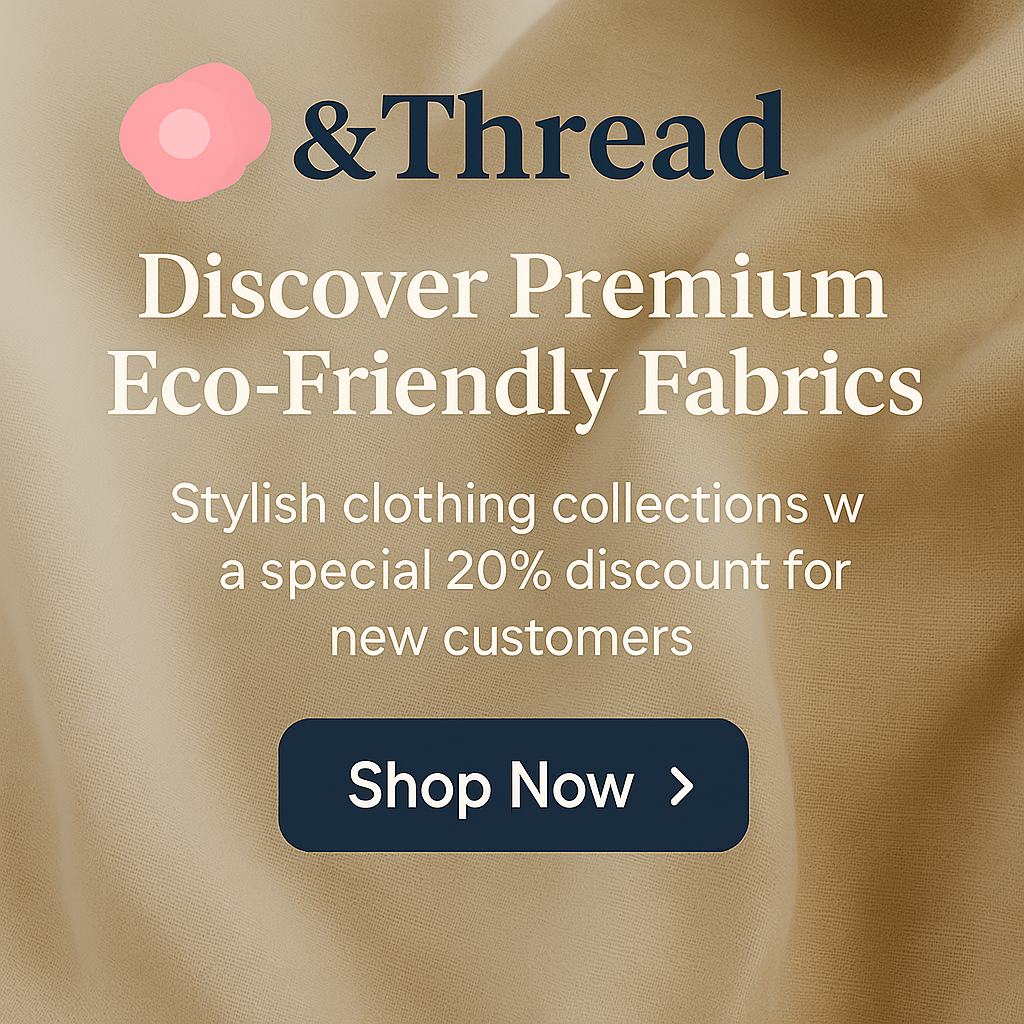}
        \caption*{\textbf{(f1)} Typo "w"}
    \end{subfigure}
    \hspace{0.01\textwidth}
    \begin{subfigure}{0.22\textwidth}
        \includegraphics[width=\linewidth]{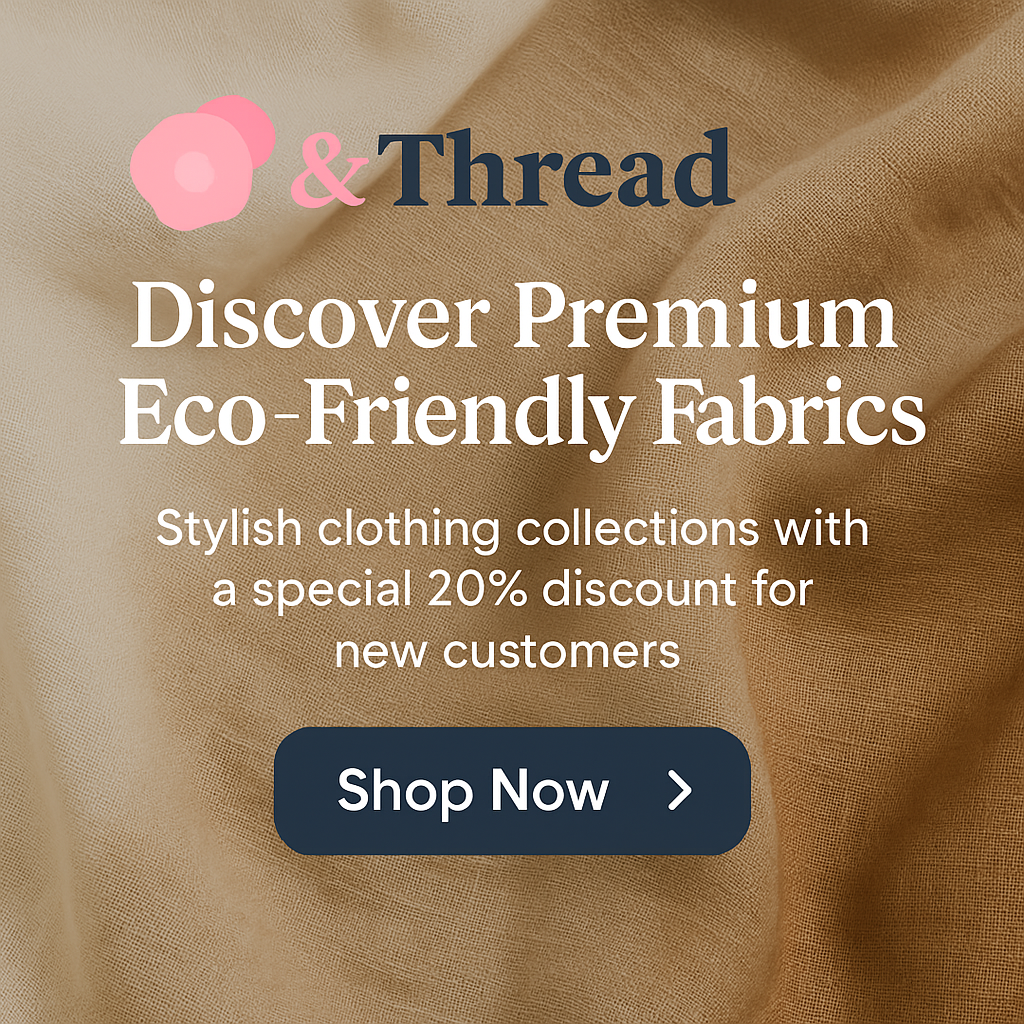}
        \caption*{\textbf{(f2)} After Refinement}
    \end{subfigure}

    \begin{subfigure}{0.22\textwidth}
        \includegraphics[width=\linewidth]{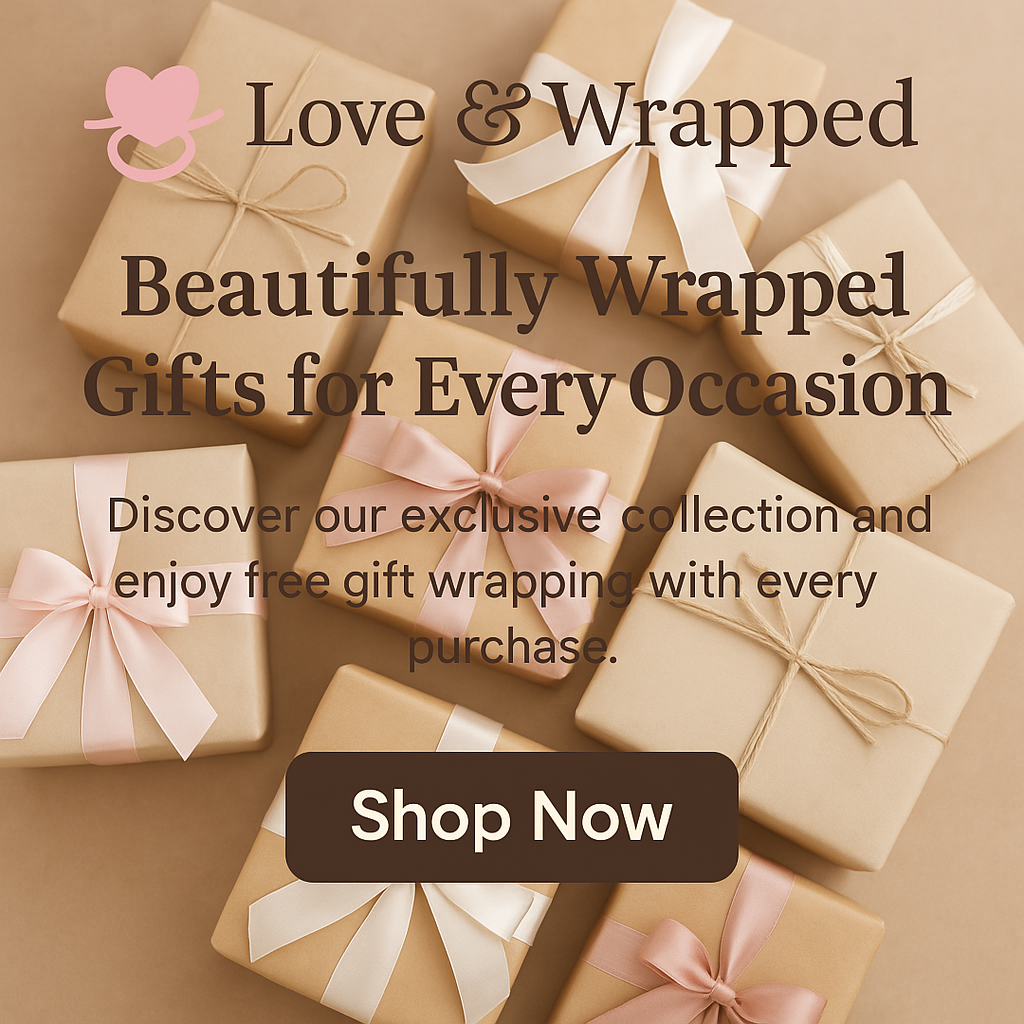}
        \caption*{\textbf{(g1)} Low Readability}
    \end{subfigure}
    \hspace{0.01\textwidth}
    \begin{subfigure}{0.22\textwidth}
        \includegraphics[width=\linewidth]{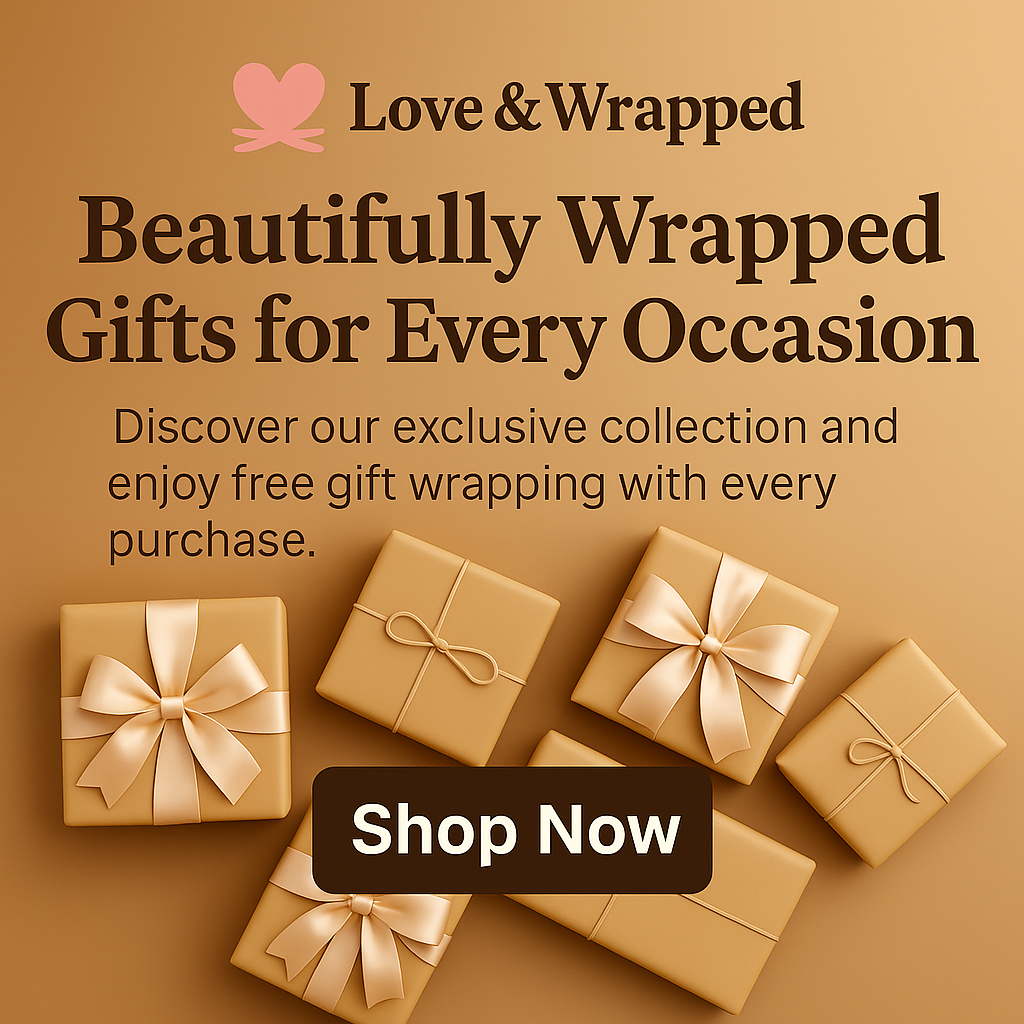}
        \caption*{\textbf{(g2)} After Refinement}
    \end{subfigure}
    \hspace{0.01\textwidth}
    \begin{subfigure}{0.22\textwidth}
        \includegraphics[width=\linewidth]{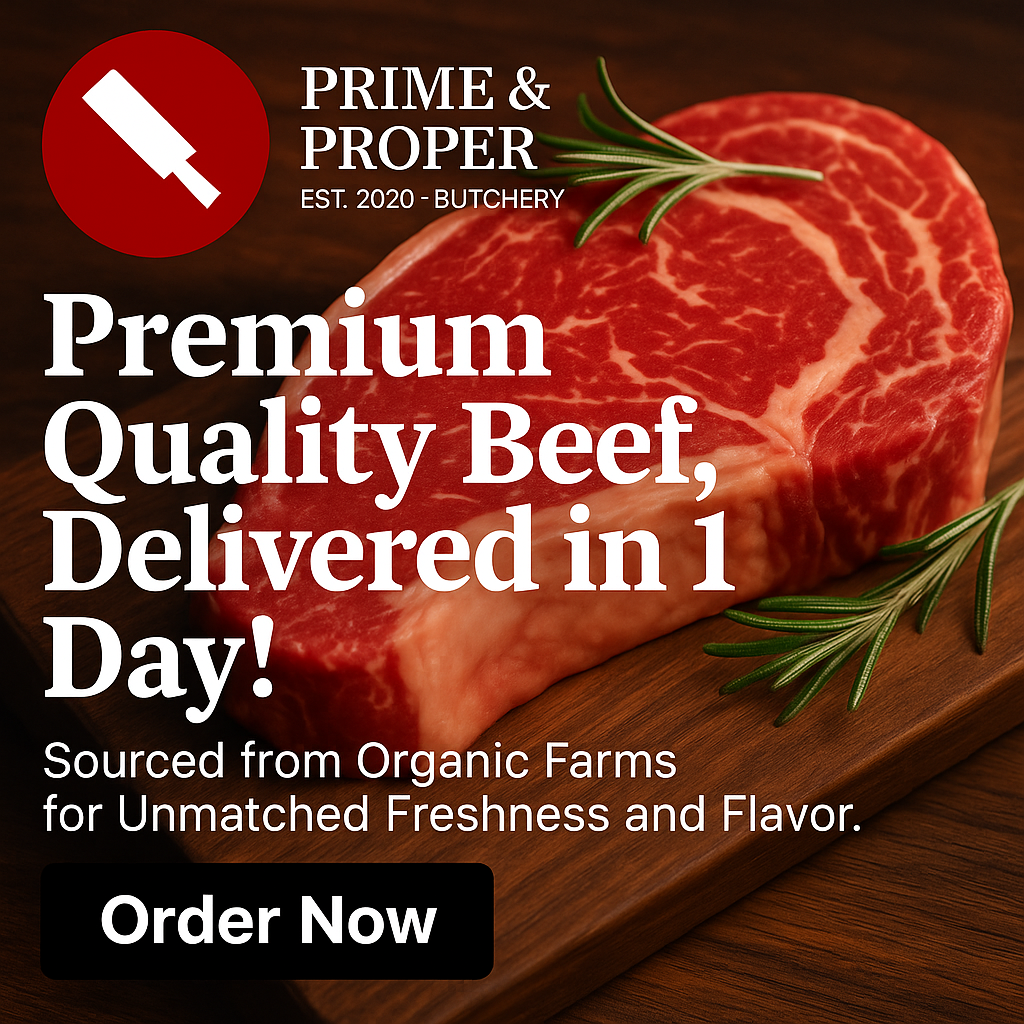}
        \caption*{\textbf{(h1)} No Emphasis}
    \end{subfigure}
    \hspace{0.01\textwidth}
    \begin{subfigure}{0.22\textwidth}
        \includegraphics[width=\linewidth]{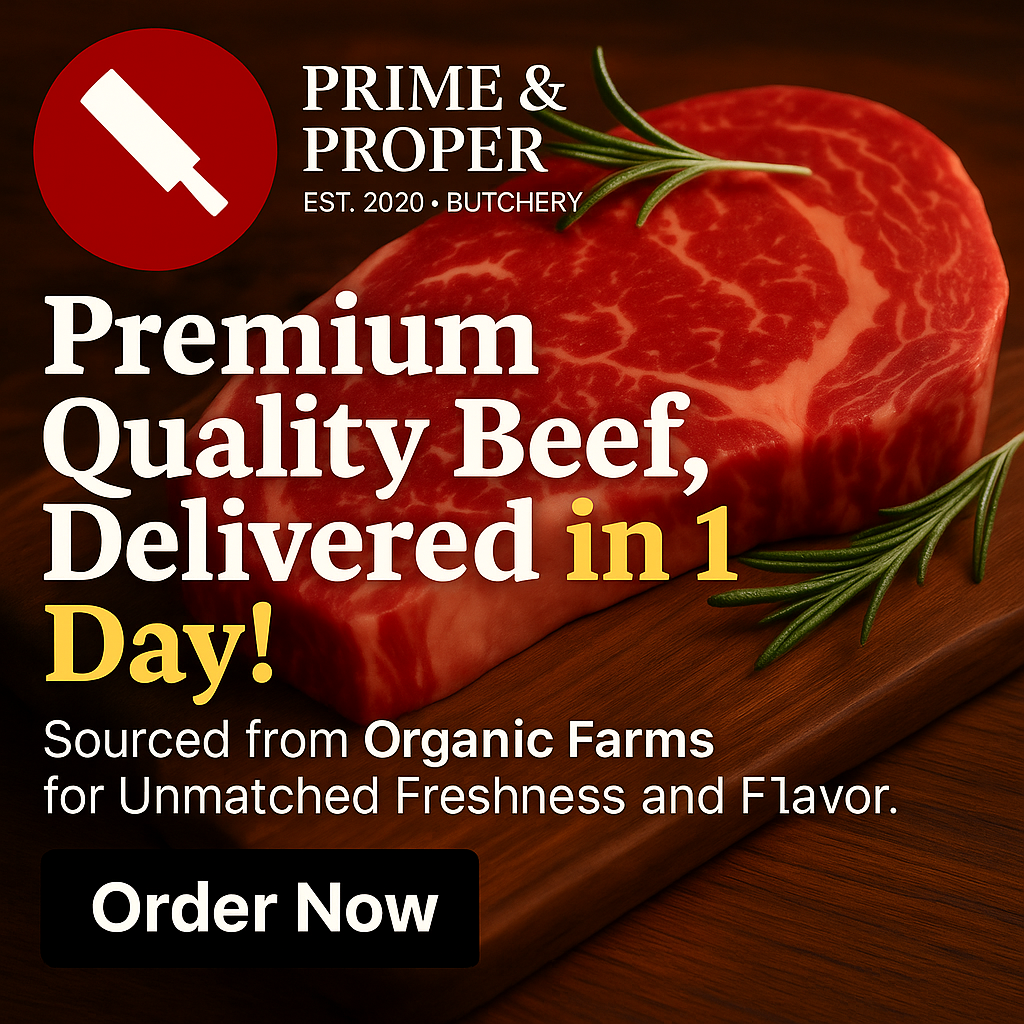}
        \caption*{\textbf{(h2)} After Refinement}
    \end{subfigure}
    
    \caption{Visualization of our system's automatic refinements. It effectively corrects subtle but important issues in ad designs, including logo placement, layout balance, background relevance, and text clarity.}
    \label{fig:mimo_core_refine}
\end{figure*}

Figure~\ref{fig:mimo_core_refine} illustrates the iterative refinement process of the \texttt{MIMO-Core} pipeline. Starting from an initial ad draft generated by the content creation team, the system undergoes multiple evaluation and revision cycles. Each round addresses specific issues—such as poor logo visibility, unbalanced layout, or weak CTA emphasis—based on textual feedback from evaluation agents. The resulting improvement highlights the effectiveness of hierarchical refinement in producing commercially viable designs.

\begin{figure*}[h]
    \centering
    \begin{subfigure}{0.22\textwidth}
        \includegraphics[width=\linewidth]{figs/logo_double_after.png}
        \caption*{\textbf{(a1)} MIMO Refined}
    \end{subfigure}
    \hspace{0.01\textwidth}
    \begin{subfigure}{0.22\textwidth}
        \includegraphics[width=\linewidth]{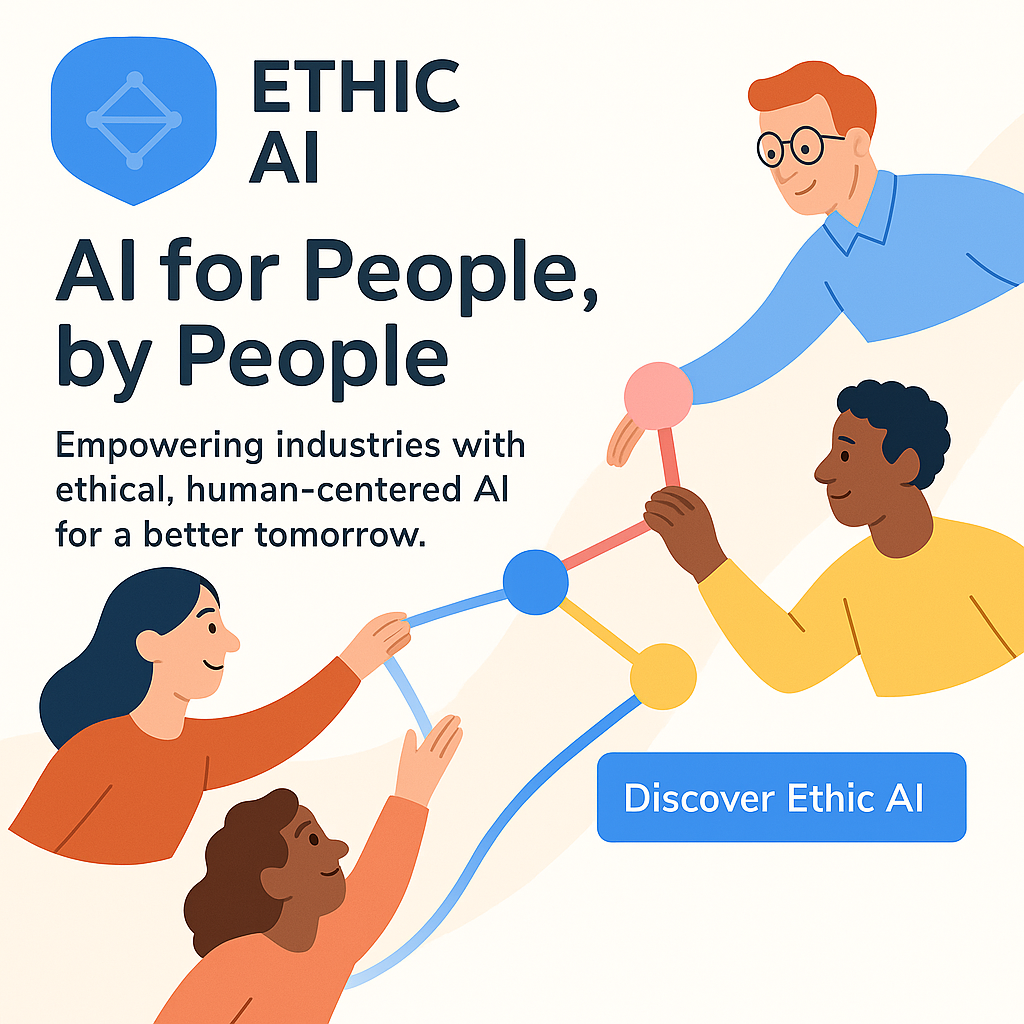}
        \caption*{\textbf{(a2)} MIMO (Style 1)}
    \end{subfigure}
    \hspace{0.01\textwidth}
    \begin{subfigure}{0.22\textwidth}
        \includegraphics[width=\linewidth]{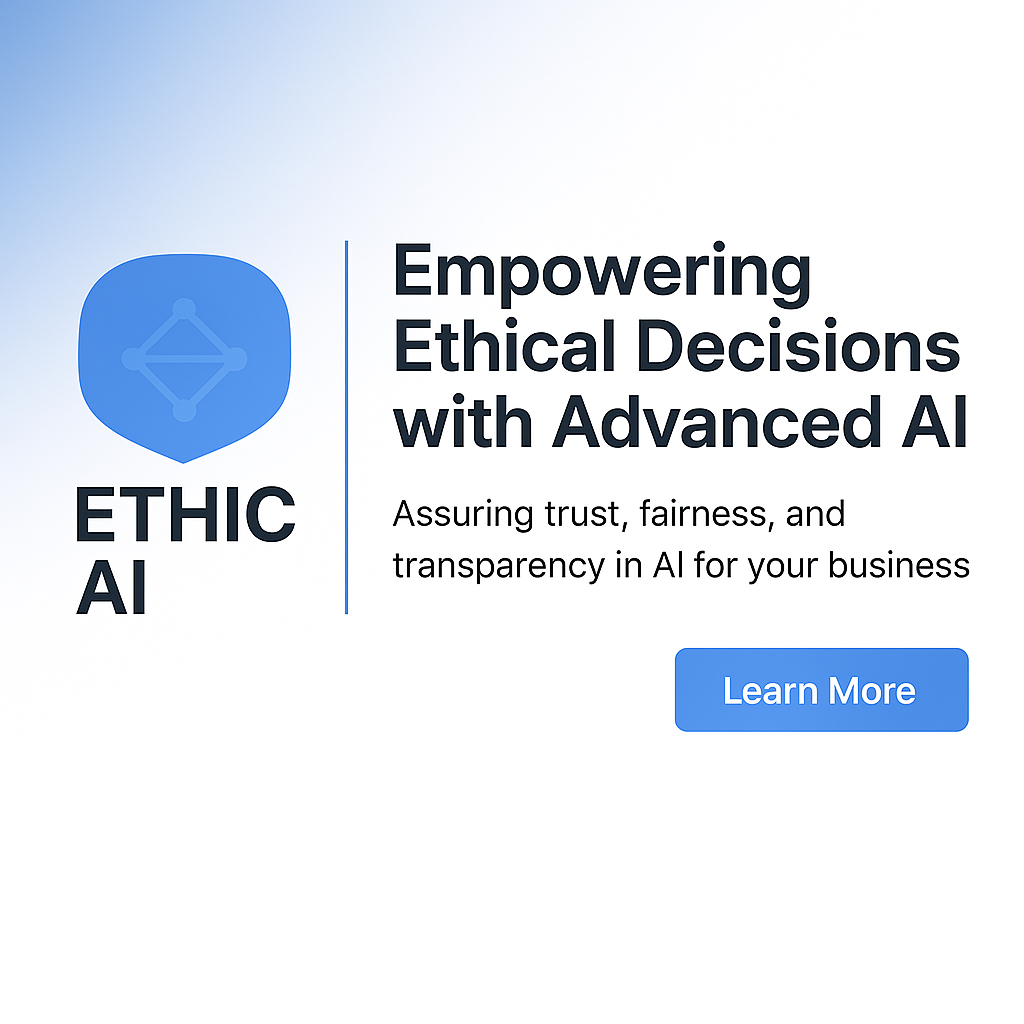}
        \caption*{\textbf{(b1)} MIMO (Style 2)}
    \end{subfigure}
    \hspace{0.01\textwidth}
    \begin{subfigure}{0.22\textwidth}
        \includegraphics[width=\linewidth]{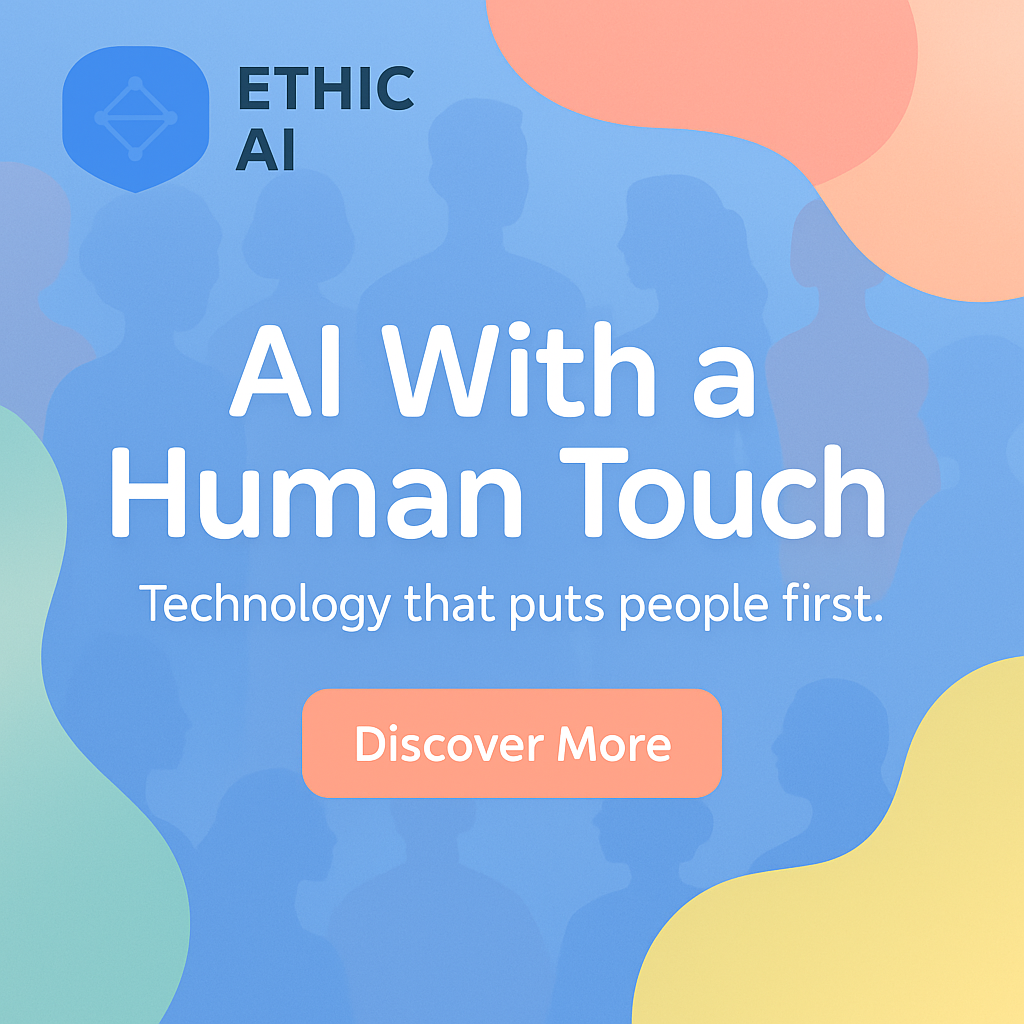}
        \caption*{\textbf{(b2)} MIMO (Style 3)}
    \end{subfigure}

    \begin{subfigure}{0.22\textwidth}
        \includegraphics[width=\linewidth]{latex/figs/layout_scale_after.png}
        \caption*{\textbf{(c1)} MIMO Refined}
    \end{subfigure}
    \hspace{0.01\textwidth}
    \begin{subfigure}{0.22\textwidth}
        \includegraphics[width=\linewidth]{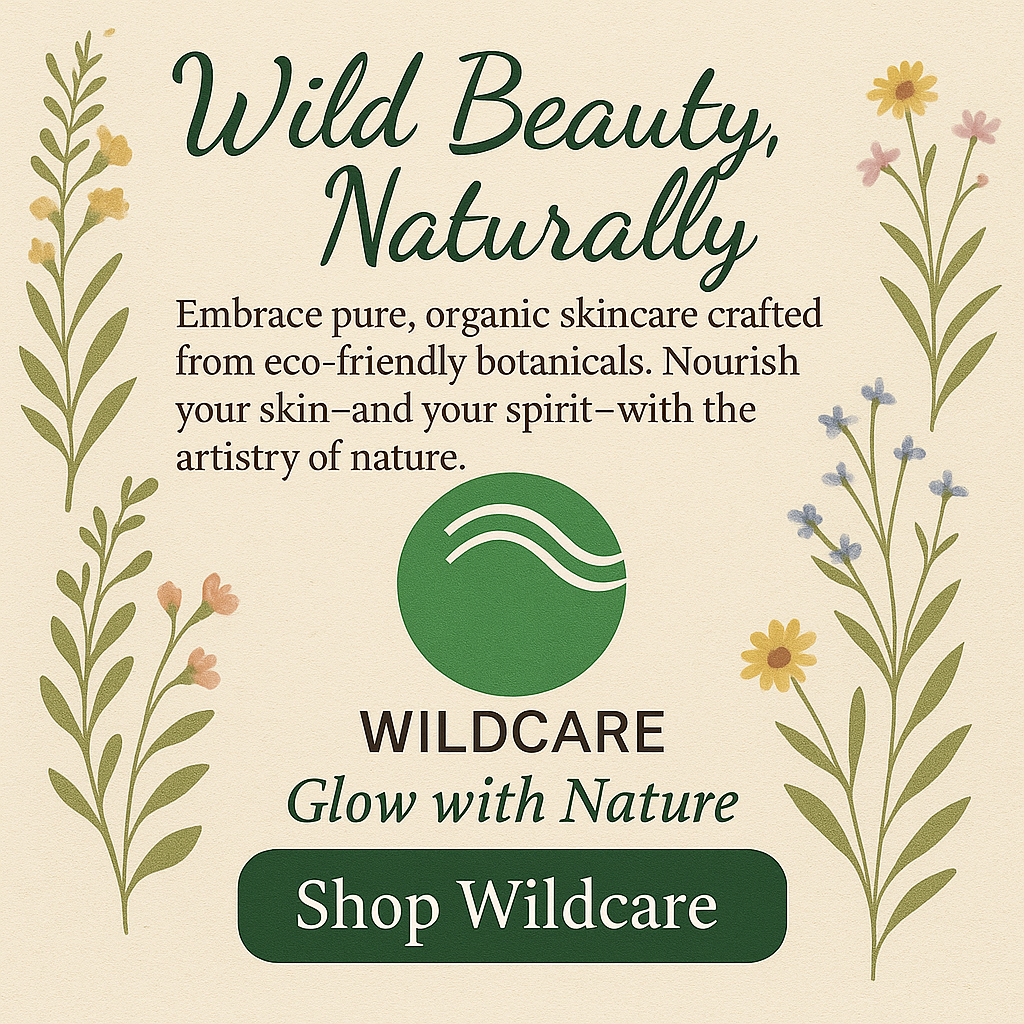}
        \caption*{\textbf{(c2)}  MIMO (Style 1)}
    \end{subfigure}
    \hspace{0.01\textwidth}
    \begin{subfigure}{0.22\textwidth}
        \includegraphics[width=\linewidth]{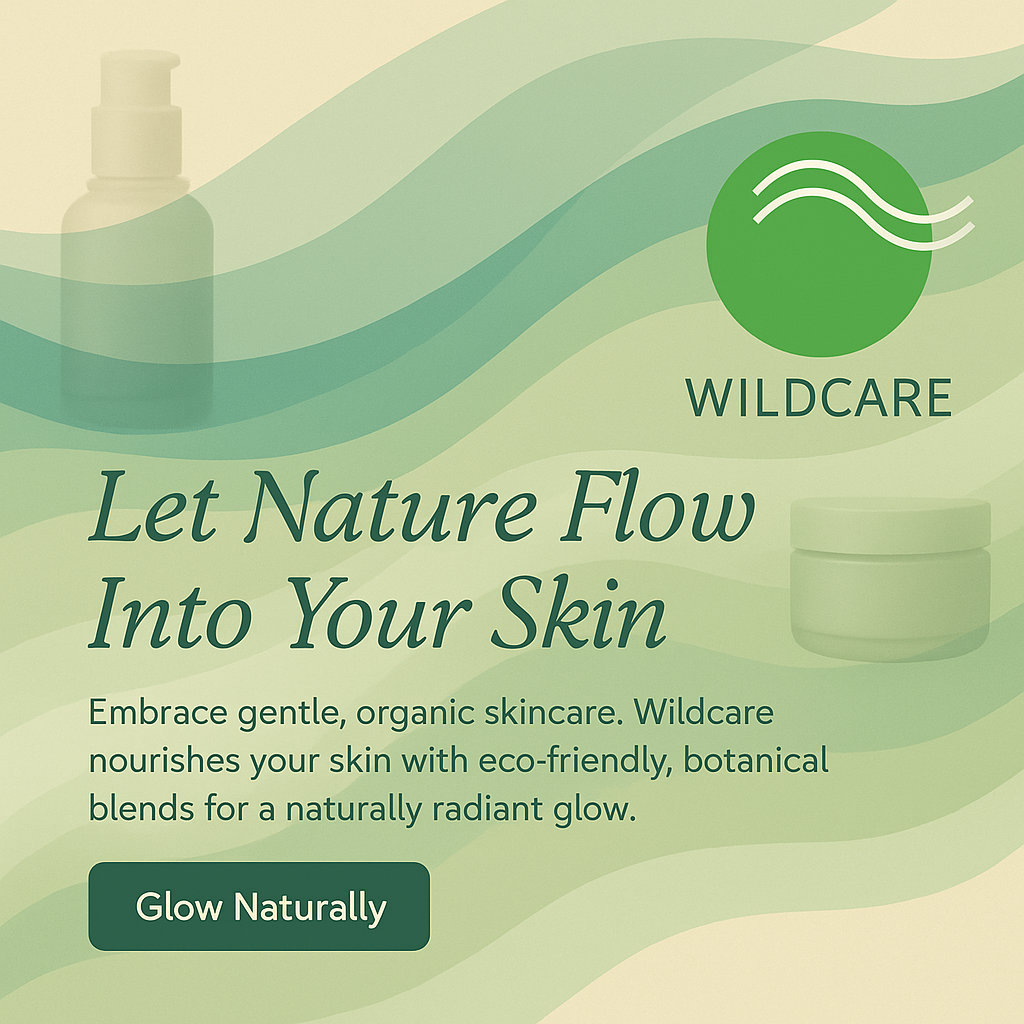}
        \caption*{\textbf{(d1)}  MIMO (Style 2)}
    \end{subfigure}
    \hspace{0.01\textwidth}
    \begin{subfigure}{0.22\textwidth}
        \includegraphics[width=\linewidth]{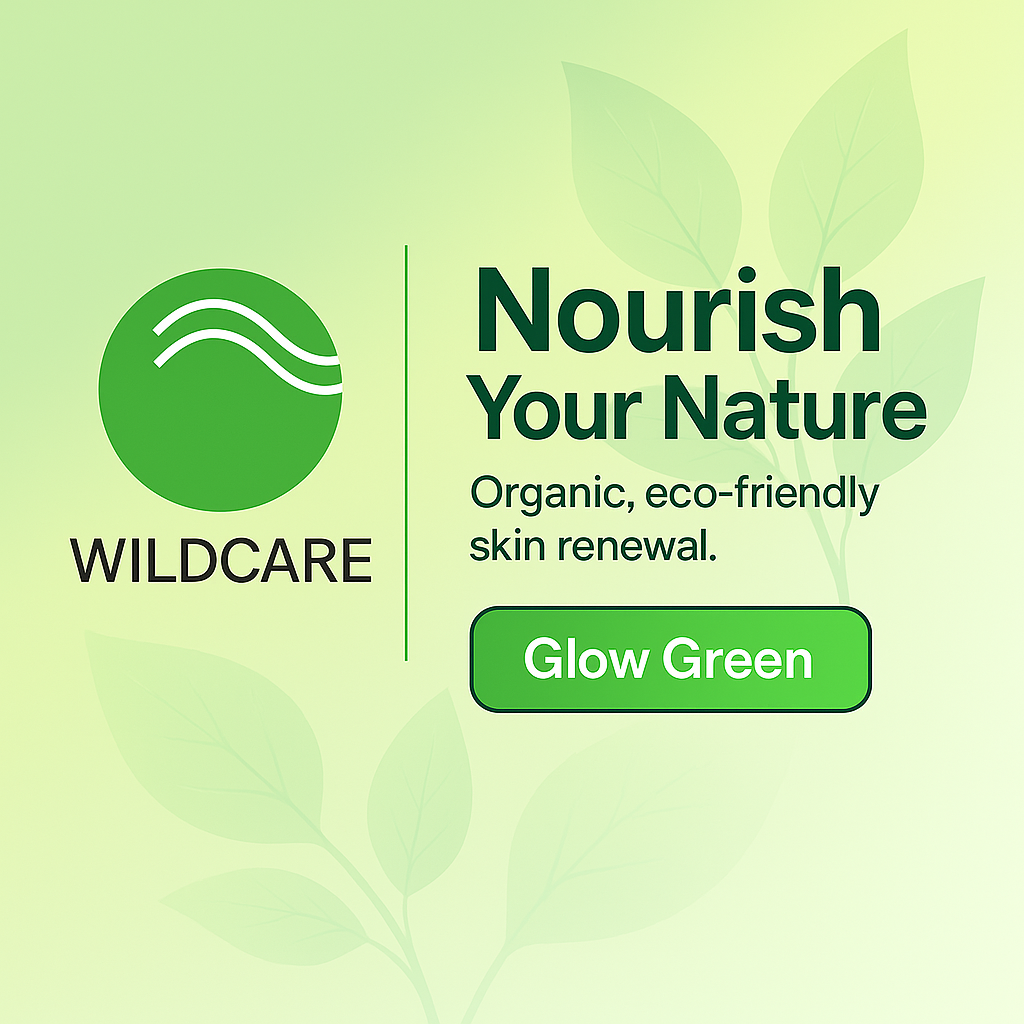}
        \caption*{\textbf{(d2)}  MIMO (Style 3)}
    \end{subfigure}

    \begin{subfigure}{0.22\textwidth}
        \includegraphics[width=\linewidth]{latex/figs/background2_after.png}
        \caption*{\textbf{(e1)} MIMO Refiend}
    \end{subfigure}
    \hspace{0.01\textwidth}
    \begin{subfigure}{0.22\textwidth}
        \includegraphics[width=\linewidth]{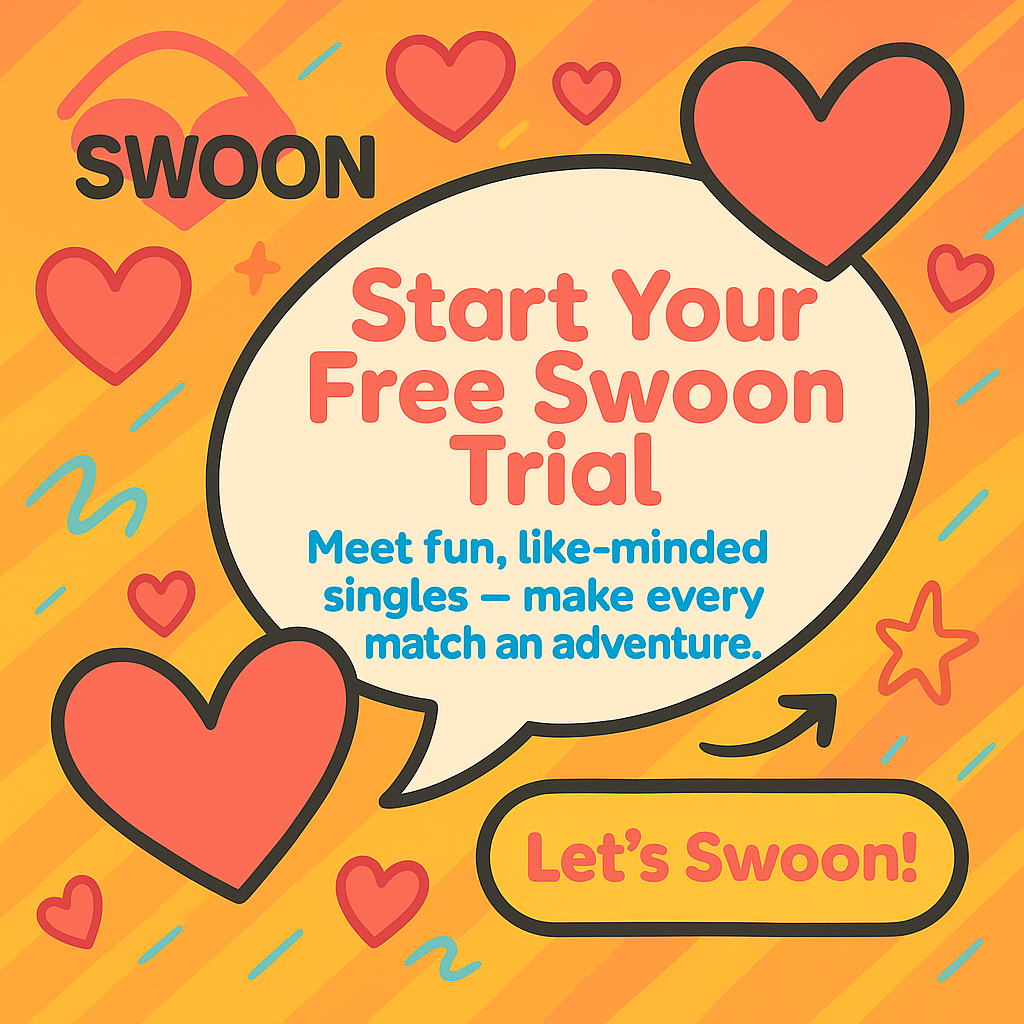}
        \caption*{\textbf{(e2)} MIMO (Style 1)}
    \end{subfigure}
    \hspace{0.01\textwidth}
    \begin{subfigure}{0.22\textwidth}
        \includegraphics[width=\linewidth]{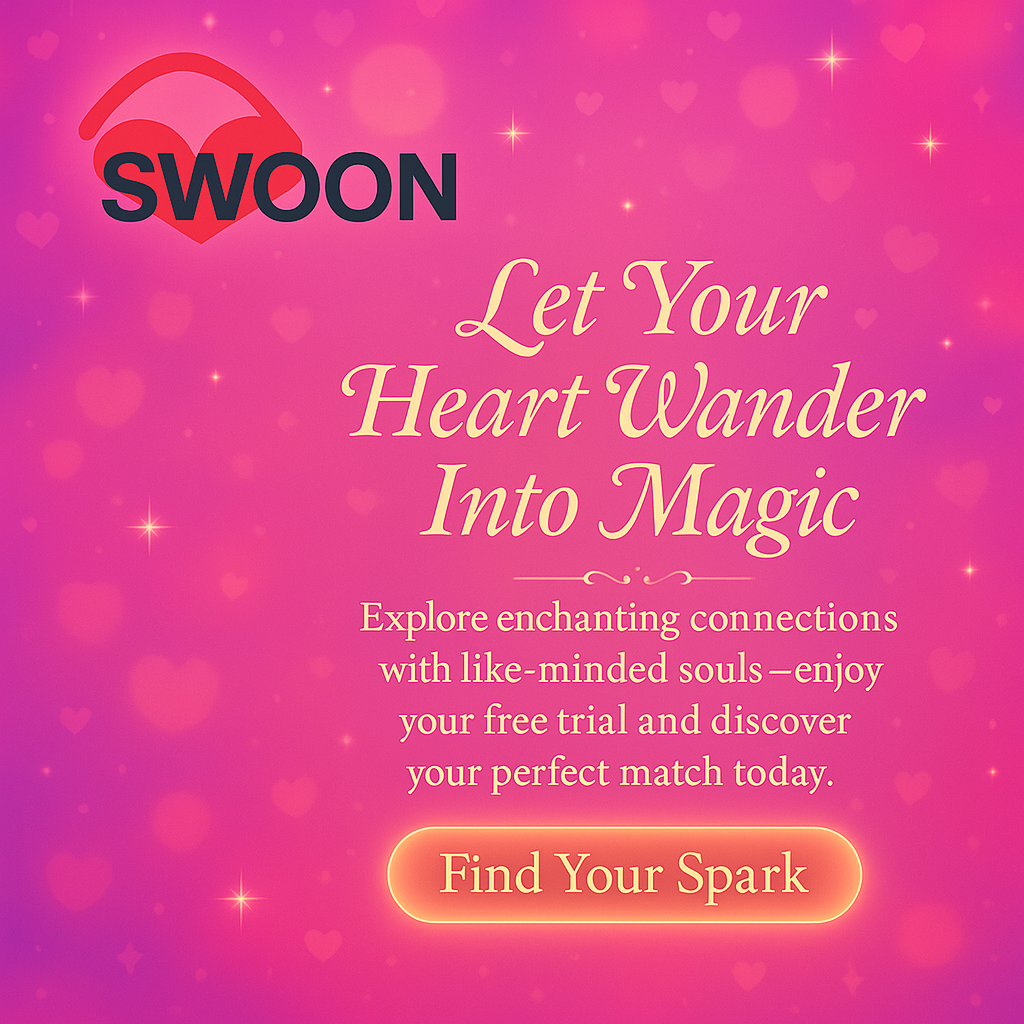}
        \caption*{\textbf{(f1)} MIMO (Style 2)}
    \end{subfigure}
    \hspace{0.01\textwidth}
    \begin{subfigure}{0.22\textwidth}
        \includegraphics[width=\linewidth]{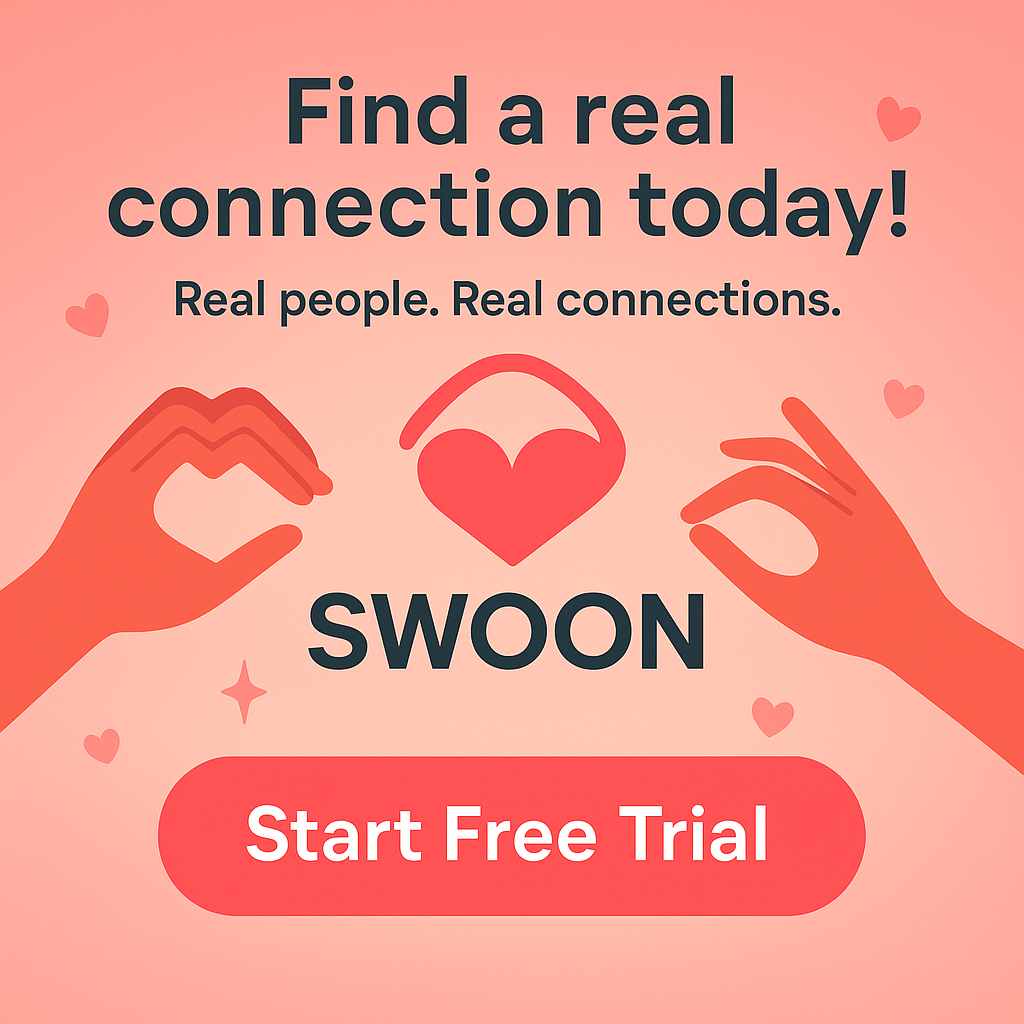}
        \caption*{\textbf{(f2)} MIMO (Style 3)}
    \end{subfigure}

    \begin{subfigure}{0.22\textwidth}
        \includegraphics[width=\linewidth]{latex/figs/low_contrast_after.png}
        \caption*{\textbf{(g1)} MIMO Refined}
    \end{subfigure}
    \hspace{0.01\textwidth}
    \begin{subfigure}{0.22\textwidth}
        \includegraphics[width=\linewidth]{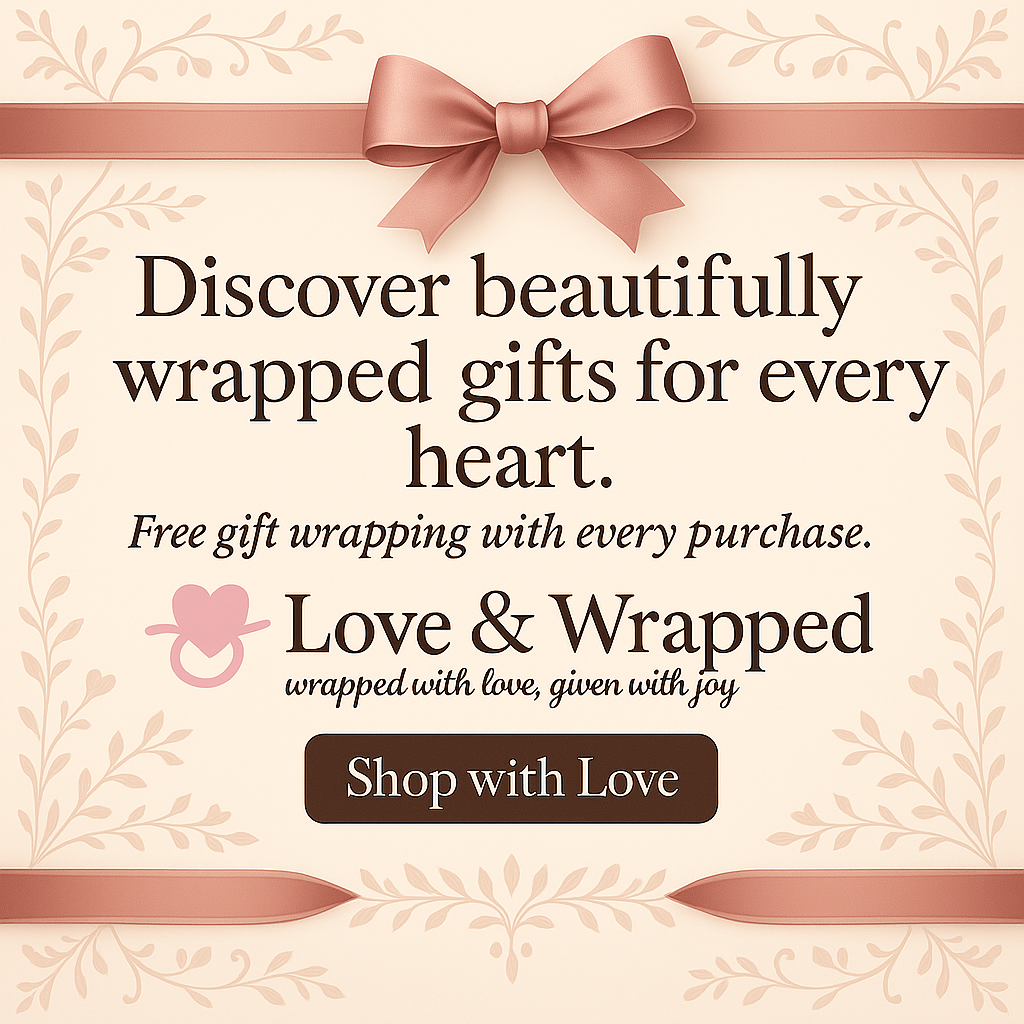}
        \caption*{\textbf{(g2)} MIMO (Style 1)}
    \end{subfigure}
    \hspace{0.01\textwidth}
    \begin{subfigure}{0.22\textwidth}
        \includegraphics[width=\linewidth]{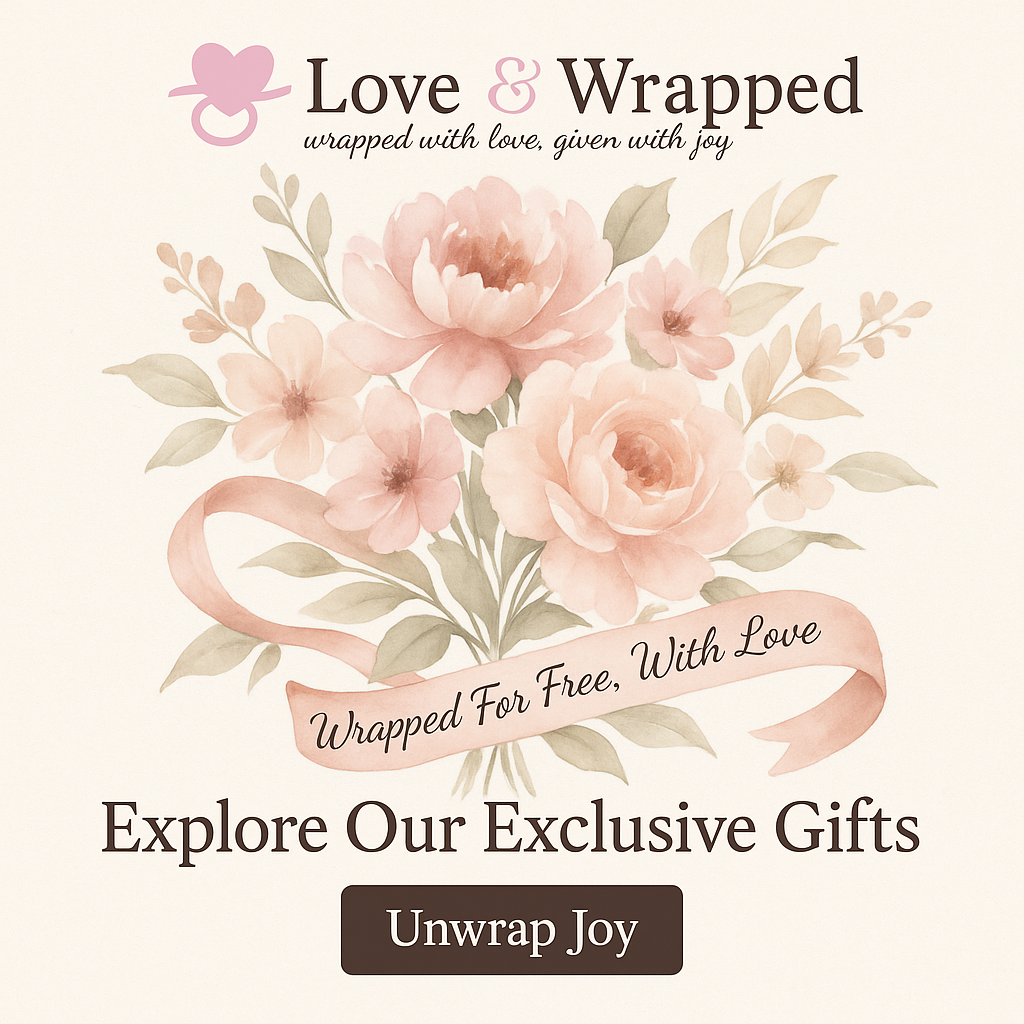}
        \caption*{\textbf{(h1)} MIMO (Style 2)}
    \end{subfigure}
    \hspace{0.01\textwidth}
    \begin{subfigure}{0.22\textwidth}
        \includegraphics[width=\linewidth]{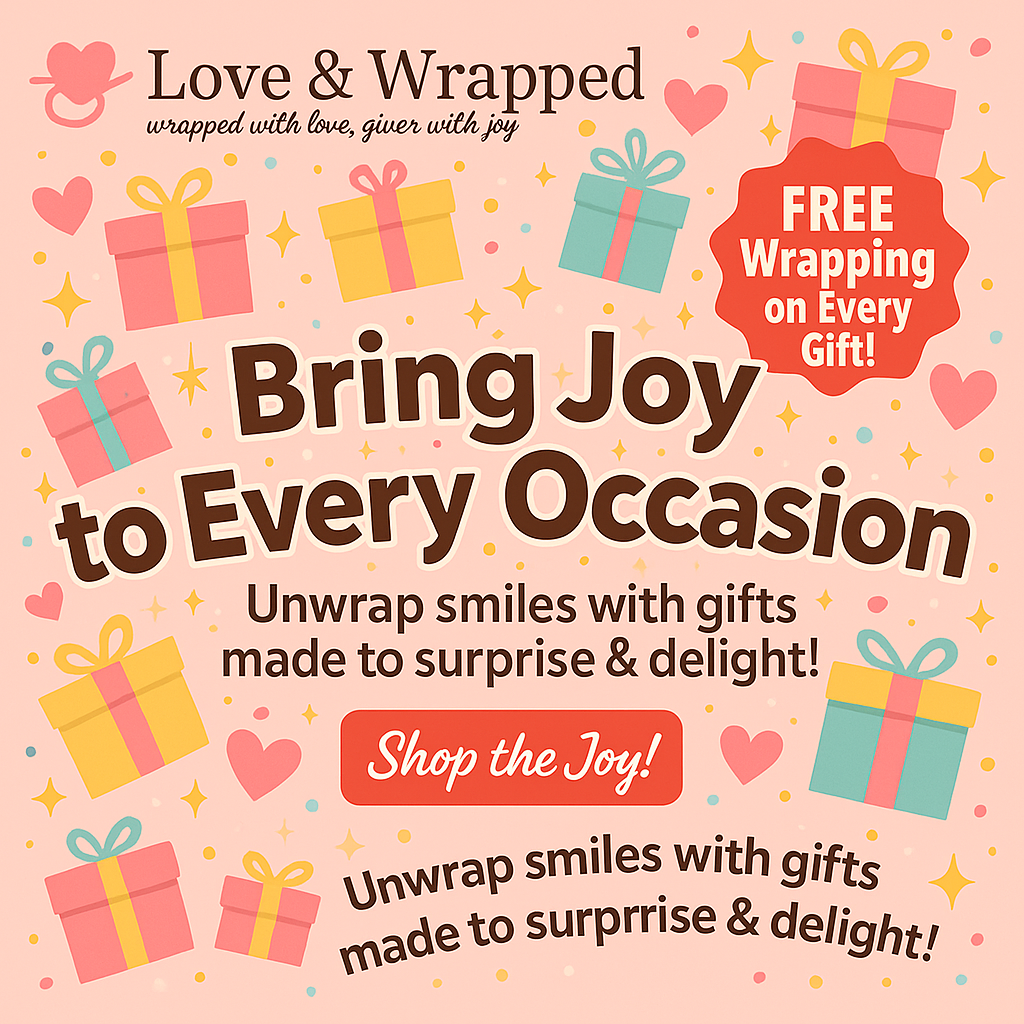}
        \caption*{\textbf{(h2)} MIMO (Style 3)}
    \end{subfigure}
    
    \caption{Visualization of MIMO-Loop with the different style of ad banners.}
    \label{fig:mimo_loop_styles}
\end{figure*}

\clearpage
Figure~\ref{fig:mimo_loop_styles} showcases the stylistic diversity generated by \texttt{MIMO-Loop}. For a single product prompt and logo, multiple design styles are instantiated in parallel, producing banners that vary in tone, layout, and visual aesthetics (e.g., minimalist vs. vibrant, modern vs. traditional). This visualization demonstrates MIMO-Loop’s ability to generate style-aligned outputs while retaining core brand identity and campaign intent. The comparison further shows how different stylistic framings can shift emphasis and design focus, providing valuable choices for downstream selection and human feedback.

\clearpage
\section{Hierarchical Multi-Agent Team Prompt for Ad Image Generation}

To support the multi-agent setting in our system, we designed a hierarchical agent framework where the task is decomposed into content creation, evaluation, and revision teams. Each team includes specialized agents responsible for different aspects of the ad generation pipeline, and the system follows a coordination protocol governed by return signals (e.g., \texttt{FINISH}, \texttt{Default}) and revision limits.

The agent teams are organized as follows:
\begin{itemize}
    \item \textbf{ContentCreationTeam}: Includes \texttt{Copywriter} and \texttt{ImageResearcher} for generating textual and visual elements.
    \item \textbf{EvaluationTeam}: Includes \texttt{TextContentEvaluator}, \texttt{LayoutEvaluator}, and \texttt{BackgroundImageEvaluator}, each focusing on a single evaluation dimension.
    \item \textbf{GraphicRevisor}: Revises the ad banner based on the structured feedback from EvaluationTeam.
\end{itemize}

The system supports a maximum of three image revisions. Once this threshold is reached, the system terminates with the return signal \texttt{"FINISH"}.

\vspace{0.5em}
\noindent Below is the full hierarchical agent prompt structure used in our implementation:

\begin{tcolorbox}[colback=gray!5!white, colframe=gray!70!black, title=Hierarchical Agent Prompt Structure, fonttitle=\bfseries, sharp corners=southwest, boxrule=0.5pt, breakable]
\scriptsize
\begin{verbatim}
{
  "team": {
    "team": "Default",
    "return": "FINISH",
    "prompt": "Make a **single** catchy square AD banner image for {item}. 
      Make sure that there is a **CTA button icon** and **logo** in the banner. 
      Furthermore, make sure that the banner stands out when small, and has a good background image. 
      If the Ad banner image has already been revised multiple times, ensure at most 3 revisions are the limitations. 
      If over 3 revisions, just return 'FINISH' and stop the process. 
      Final output MUST be in the form of: {\"images\": [...], \"instructions\": [...], \"current_content\": [...]}.",
      
    "additional_prompt": "Ensure the AD banner is visually appealing and the text is persuasive. 
      You must evaluate the image before reporting back to FINISH. 
      Furthermore, make sure to remind the evaluators that there is a revised image whenever there is one.",

    "ContentCreationTeam": {
      "team": "ContentCreation",
      "return": "Default",
      "prompt": "Create the text content and find high-quality images for the product we are advertising. 
        You only need to generate 1 image and the text to go with it. 
        If there is a reference image, you need to pass the path to Image Researcher.",
      "additional_prompt": "Ensure the texts you created and images are cohesive and high-quality, 
        and the path to the image is properly returned.",

      "Copywriter": {
        "prompt": "You are a Copywriter responsible for creating the text content for the AD banner, 
          including the headline, subheadline, and CTA text. 
          You can use tools to find information about the product. 
          CTA text must be short, catchy and concise. 
          Do not generate untrue, misleading or incorrect information.",
        "tools": [4]
      },

      "ImageResearcher": {
        "prompt": "You are an Image Researcher responsible for finding high-quality images of the product. 
          Follow the instructions from the supervisor. 
          You need to pass detailed ad image generation request to the tool. 
          Only generate 1 image at a time. 
          Only 1 logo in the image. 
          All text must be high contrast and visible.",
        "tools": [4, 12]
      }
    },

    "EvaluationTeam": {
      "team": "Evaluation",
      "return": "Default",
      "prompt": "Review the text content and image design of the AD banner for clarity, persuasiveness, correctness, and visual appeal. 
        You must evaluate the text and image separately. 
        If nothing is wrong, say 'No changes needed'. 
        If over 3 revisions, return 'FINISH'.",
      "additional_prompt": "Check whether the image is revised. 
        When given a revised image path, evaluate again.",

      "TextContentEvaluator": {
        "prompt": "You are a Text Content Evaluator responsible for reviewing the text for clarity, persuasiveness, and correctness. 
          Make sure the text is **visible and readable**. 
          Must use tools to check the text rendered in the image. 
          Only focus on text (not layout/background).",
        "tools": [4, 12]
      },

      "LayoutEvaluator": {
        "prompt": "You are a Layout Evaluator responsible for reviewing element placement and overall visual flow. 
          Give layout-specific suggestions only (not text or background).",
        "tools": [4, 12]
      },

      "BackgroundImageEvaluator": {
        "prompt": "You are an Image Evaluator responsible for judging if the background is suitable. 
          Give suggestions on better imagery if needed. 
          Only focus on background (not text/logo).",
        "tools": [4, 12]
      }
    },

    "GraphicRevisor": {
      "prompt": "You are a Graphic Revisor responsible for modifying the AD image per evaluation feedback. 
        Use the most recently generated image. 
        Give clear instructions to the editing tool.",
      "tools": [4, 12]
    }
  },
  "intermediate_output_desc": "Dictionary format.",
  "int_out_format": "Dictionary"
}
\end{verbatim}
\end{tcolorbox}

\clearpage
\section{Inpainting for Logo and Pruduct Image Replacement}
\label{logo}
\begin{figure}[h]
    \centering
    \includegraphics[width=\textwidth]{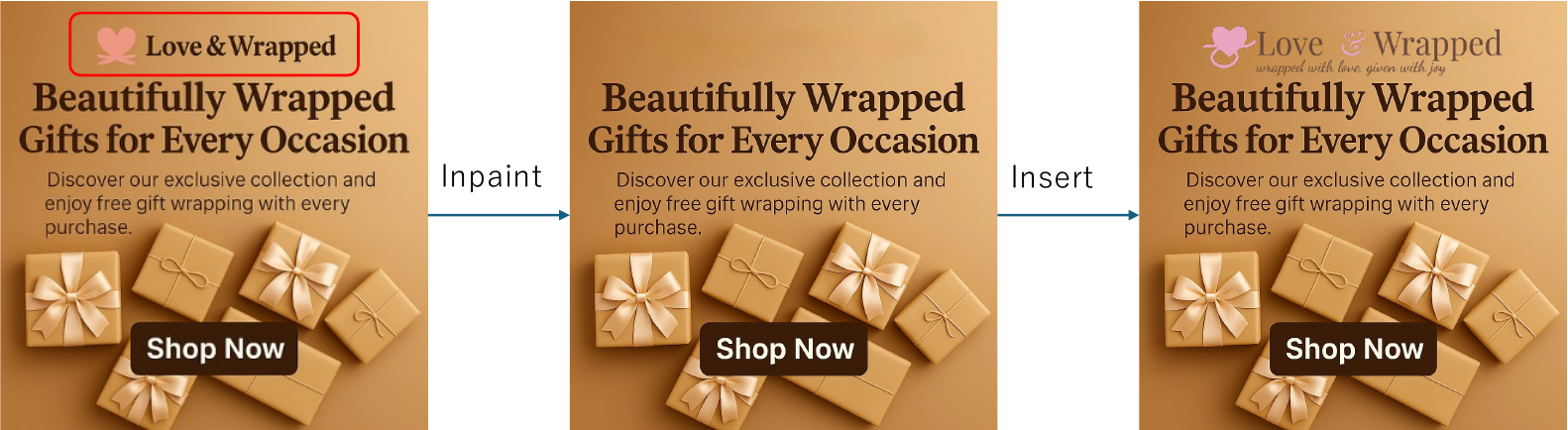}
    \caption{Inpainting example for logo replacement.}
    \label{fig:inpaint}
\end{figure}

In this section,  we describe how to post-process the generated banner image for the logo and product image inpainting to guarantee the logo's correctness.

For a generated banner image, while we input the logo and product image for generation, due to the nature of generative models, the correctness of them in the banner image is not guaranteed.
This creates a fatal weakness in using generative models for real-world business banner image generation, as the logo generated through iterations may not accurately capture the details of the original logo and product image.
However, for a banner image, a wrong logo and product image banner is certainly not usable.
To address this issue, we use inpaint to do the replacement to ensure the their correctness.

We show an example in the Figure \ref{fig:inpaint}.
For this figure, the first generated final banner image has a wrong logo which is not the same with the actual one.
We first ask an inpaint model to inpaint the original logo to fit the background image.
After that, we insert the original transparent logo at the same inpaint area so that the size can fit with the original logo design generated by the MIMO method.
In this way, we are able to replace the generated logo and product image with the actual ones while ensuring the original design from the MIMO is kept.

\section{Token and Cost Analysis}
\label{app:token}

We analyze the token usage and cost of our MIMO framework compared to a GPT-4o baseline, based on the latest Azure OpenAI pricing for GPT-4o’s multimodal model. Text input and output tokens are billed at \$0.00004 and \$0.00008 per token, respectively. For image generation, GPT-4o consumes approximately 1,105 output tokens per image at 512×512 resolution and medium detail, resulting in a per-image cost of \$0.0884.

\textbf{GPT-4o Baseline} performs one-shot banner generation using approximately 1,000 text tokens (500 input, 500 output) and one image, for a total cost of:
\begin{itemize}
    \item Text: (500 × \$0.00004) + (500 × \$0.00008) = \$0.06
    \item Image: \$0.0884
    \item \textbf{Total: \$0.15}
\end{itemize}

\textbf{MIMO-Core$^-$} (a simplified one-round version) uses about 4,000 text tokens (2,000 input, 2,000 output) and one image:
\begin{itemize}
    \item Text: (2,000 × \$0.00004) + (2,000 × \$0.00008) = \$0.24
    \item Image: \$0.0884
    \item \textbf{Total: \$0.33}
\end{itemize}

\textbf{MIMO-Core} includes three rounds of generation, evaluation, and refinement, totaling approximately 10,800 text tokens (5,400 input, 5,400 output) and three images:
\begin{itemize}
    \item Text: (5,400 × \$0.00004) + (5,400 × \$0.00008) = \$0.65
    \item Image: 3 × \$0.0884 = \$0.265
    \item \textbf{Total: \$0.91}
\end{itemize}

\textbf{MIMO-Core + Loop}, which runs three parallel MIMO-Core branches, a judging phase (five agents × three candidates), and one final refinement, consumes about 42,000 text tokens (21,000 input, 21,000 output) and four images:
\begin{itemize}
    \item Text: (21,000 × \$0.00004) + (21,000 × \$0.00008) = \$2.52
    \item Image: 4 × \$0.0884 = \$0.3536
    \item \textbf{Total: \$2.87}
\end{itemize}

While MIMO introduces higher computational costs (ranging from 2–20× over the baseline), it consistently yields higher-quality banners, stronger visual fidelity, and more robust multi-agent evaluations—justifying its use in high-stakes advertising scenarios.

\clearpage
\section{LLM Evaluation Prompt}
\label{app:eval_prompt}

To enable structured and consistent assessments of banner quality, we follow~\cite{wang2025banneragency} a unified evaluation prompt for LLM-based comparison. Given two ad images from the same campaign, the model is instructed to rate and justify each image across six critical design dimensions: audience alignment, logo visibility, CTA effectiveness, copywriting quality, brand consistency, and overall aesthetics. Each criterion is scored on a 0–5 scale with detailed reasoning, and the final judgment is returned in a standardized JSON format for programmatic evaluation.

\begin{tcolorbox}[colback=gray!5!white, colframe=gray!70!black, title=LLM Evaluation Prompt, fonttitle=\bfseries, sharp corners=southwest, boxrule=0.5pt]
\scriptsize
\ttfamily
Compare two ad images that are from the same ad campaign. You must choose which is better. Explain why the first or second ad image is better using the following six criteria. For each metric, rate the first and second image on a scale from 0 (very poor) to 5 (excellent), and explain the reasoning for each rating.

The ratings should be based on the following scale:\\
\ \ "0": "Very poor – The element is missing, unreadable, or completely fails in its purpose. For example, no CTA, broken layout, or irrelevant audience targeting."\\
\ \ "1": "Poor – The element exists but is confusing, hard to notice, or seriously misaligned with goals. For instance, logo is tiny and placed in a non-visible area, or the message is unclear."\\
\ \ "2": "Fair – The element is somewhat effective but has noticeable problems. For example, the CTA is present but lacks emphasis, or the visuals are cluttered or off-brand."\\
\ \ "3": "Good – The element is mostly effective with minor issues. For example, the layout works but the font size is slightly small, or the copy is clear but could be more persuasive."\\
\ \ "4": "Very good – The element is clear, aligned with the goal, and visually sound. It may need very small refinements, such as slight enlargement of text or better spacing."\\
\ \ "5": "Excellent – The element is perfect in both design and function. No changes are needed. It meets its communication goal effectively and aligns with professional visual standards."

Return your response in the following JSON format:

\{\ \\
\ \ \ "TAA": \{ "image\_1\_score": 0, "image\_1\_reason": "Your explanation here.", "image\_2\_score": 0, "image\_2\_reason": "Your explanation here." \},\\
\ \ \ "LPS": \{ "image\_1\_score": 0, "image\_1\_reason": "Your explanation here.", "image\_2\_score": 0, "image\_2\_reason": "Your explanation here." \},\\
\ \ \ "CTAE": \{ "image\_1\_score": 0, "image\_1\_reason": "Your explanation here.", "image\_2\_score": 0, "image\_2\_reason": "Your explanation here." \},\\
\ \ \ "CPYQ": \{ "image\_1\_score": 0, "image\_1\_reason": "Your explanation here.", "image\_2\_score": 0, "image\_2\_reason": "Your explanation here." \},\\
\ \ \ "BIS": \{ "image\_1\_score": 0, "image\_1\_reason": "Your explanation here.", "image\_2\_score": 0, "image\_2\_reason": "Your explanation here." \},\\
\ \ \ "AQS": \{ "image\_1\_score": 0, "image\_1\_reason": "Your explanation here.", "image\_2\_score": 0, "image\_2\_reason": "Your explanation here." \}\\
\}

Evaluation Metric Descriptions:

- TAA (Target Audience Alignment): How well the ad aligns with the theme, audience, and purpose.

- LPS (Logo Placement Score): Visibility, size, and placement of the logo.

- CTAE (Call-to-Action Effectiveness): Clarity and visual emphasis of the CTA.

- CPYQ (Copy Quality): Clarity, grammar, and persuasiveness of the headline and other text.

- BIS (Brand Integration Score): Visual and stylistic alignment with the brand identity (beyond logo).

- AQS (Aesthetic Quality Score): Overall visual appeal, color harmony, balance, and typography.
\end{tcolorbox}

In addition to pairwise LLM-based comparisons, we design a six-way evaluation protocol tailored to our structured ad design assessment needs. This prompt asks the model to evaluate six ad images from the same campaign across six granular criteria: logo placement, emphasis on key information, layout, typography, text readability, and background relevance. Each metric is scored on a scale from 1 (unacceptable) to 5 (commercial-ready), with justifications provided for every image and criterion. This comprehensive and fine-grained evaluation aligns with our goal of supporting scalable SSA (Sponsored Search Advertising) design diagnostics and quality control at the campaign level.

\begin{tcolorbox}[colback=gray!5!white, colframe=gray!70!black, 
  title=Six-Way Multi-Criteria Evaluation Prompt, fonttitle=\bfseries, 
  sharp corners=southwest, boxrule=0.5pt, breakable]
\scriptsize
\ttfamily
Compare six ad images that are from the same ad campaign.

You must assign scores to all six images using the following six criteria:

1. \textbf{Logo Placement \& Contrast (LPC)} — Checks logo visibility, sizing, and whether it contrasts with the background.
2. \textbf{Emphasis on Key Information (EKI)} — Assesses if key content like discounts or product names are visually emphasized.
3. \textbf{Layout (LAY)} — Evaluates the structure, spacing, and alignment of visual elements.
4. \textbf{Typography (TYP)} — Considers font clarity, consistency, hierarchy, and branding alignment.
5. \textbf{Text Readability \& Attractiveness (TRA)} — Rates clarity and aesthetic of text given contrast and composition.

\vspace{0.5em}
\textbf{Scoring Guidelines:}
\begin{itemize}
\item 1 – Critical error: unusable in commercial settings (e.g., unreadable text, missing logo).
\item 3 – Moderate issues: usable with revisions (e.g., poor layout or weak visual emphasis).
\item 5 – Strong: commercially ready with clean, clear, and goal-aligned design.
\end{itemize}

Provide your response in JSON format, assigning a \texttt{score} and a \texttt{reason} for each image per criterion.
\end{tcolorbox}

\section{Single-Agent Prompt Used in Ablation Study}
\label{app:single_agent_prompt}

The following prompt was used in the single-agent setting during the ablation study (Section~4.X). It compresses all responsibilities into a single instruction and serves as a baseline to evaluate the benefit of role decomposition in our multi-agent design.

\begin{tcolorbox}[colback=gray!5!white, colframe=gray!70!black, title=Single-Agent Prompt for Ad Image Generation, fonttitle=\bfseries, sharp corners=southwest, boxrule=0.5pt]
\ttfamily
I need you to make a \textbf{single} catchy square AD banner image for \{item\}. \\

Make sure that there is a \textbf{CTA button icon} and \textbf{logo} in the banner. Furthermore, make sure that the banner stands out when small, and has a good background image. \\
Ensure the AD banner is visually appealing and the text is persuasive. \\

Creating the text content for the AD banner, including the headline, subheadline, and CTA text. CTA text used in CTA button must be short, catchy, and concise. \\
Do not generate untrue, misleading, or incorrect information. \\

Only generate 1 image at a time. Only generate 1 logo in the image. All text (including logo text, CTA text, text in the image, etc.) in the image needs to be high contrast and visible. \\

Ensure the text content is clear, persuasive, and correct. Make sure the text in the image is \textbf{visible and readable} with no typos. \\
Text must be high-contrast enough to the background image. \\

Review the layout of the AD banner for proper placement of elements and overall effectiveness for an AD image. \\

The background image needs to be suitable. Make sure to give pointers to what you think is a suitable background image, if any adjustments are required.
\end{tcolorbox}

\section{Style Prompting}
\label{sec:style_prompting}

To enable diverse and brand-aligned banner generation, we introduce a style prompting mechanism as the first stage of the \texttt{MIMO-Loop} framework. This stage leverages an LLM to infer multiple stylistic directions given the advertiser’s prompt and logo.

\vspace{0.5em}
\noindent\textbf{Stylized Prompt Set.}
Given a logo–prompt pair \( (\ell, u) \), a language model generates \( k \) stylistic descriptions:
\[
\mathcal{S} = \{s_1, s_2, \dots, s_k\}
\]
where each style prompt \( s_i \) is a short textual description representing a unique visual theme (e.g., minimalist, futuristic, or elegant). These styles are conditioned on logo characteristics (e.g., shape, color palette) and prompt semantics (e.g., product category, tone).

\vspace{0.5em}
\noindent\textbf{Style Selection.}
Out of the \( k \) generated styles, a style selection agent chooses a subset \( \mathcal{S}^* \) of \( n \) promising candidates:
\[
\mathcal{S}^* = \{s_{i_1}, s_{i_2}, \dots, s_{i_n}\} \subseteq \mathcal{S}
\]
The selection is based on compatibility between the prompt–logo pair \( (\ell, u) \) and each stylistic candidate \( s_i \), aiming to preserve brand alignment while encouraging stylistic diversity.

\vspace{0.5em}
\noindent\textbf{Style Prompting Instruction.}
The instruction used to query the LLM for diverse styles is provided below.

\begin{tcolorbox}[colback=gray!5!white, colframe=gray!70!black, 
  title=Style Prompting Instruction, fonttitle=\bfseries, 
  sharp corners=southwest, boxrule=0.5pt, breakable]
\ttfamily
Given the following product information and brand logo description, generate \textbf{five distinct style prompts} that describe alternative visual directions for designing an ad banner.

Each style prompt should be:
\begin{itemize}
  \item A concise one-sentence description (10–20 words).
  \item Visually grounded: refer to layout, mood, color palette, or design theme.
  \item Coherent with the brand tone and product characteristics.
  \item Diverse: avoid overlapping ideas or redundant phrasing across styles.
\end{itemize}

Use the product information and logo cues to infer appropriate stylistic themes. For example, if the logo is elegant and monochrome, one style could be “minimalist black-and-white luxury layout”; if the product is tech-related, another might be “futuristic neon over dark background.”

\vspace{0.3em}
\textbf{Product Description:} [Insert here] \\
\textbf{Logo Description:} [Insert here]

\vspace{0.5em}
Return your response as:
\{
  "style\_1": "...",
  "style\_2": "...",
  "style\_3": "...",
  "style\_4": "...",
  "style\_5": "..."
\}
\end{tcolorbox}

\section{Baseline prompt design}
\label{basepromptdesign}
Text-to-image generative models often require carefully designed prompts to produce high-quality outputs. To bridge the gap between advertising-specific prompts and the formats expected by different generative models, we implemented an agent, T2IPromptGenerator, which automatically reformulates advertising prompts into model-compatible inputs. The agent operates in a few-shot manner, guided by a system prompt that provides instructions along with several examples of the desired prompt format to align outputs with user intent.

\begin{tcolorbox}[colback=gray!5!white, colframe=gray!70!black, title=System Prompt for T2IPromptGenerator, fonttitle=\bfseries, sharp corners=southwest, boxrule=0.5pt]
\ttfamily
You are an image prompt generator for web advertisement. Given the user's request about the image or explanation about their product, you will generate a detailed prompt for the image generation API.

When generating the prompt, make sure:

- There is a **CTA button icon** and **logo** in the banner. \\
- Include text contents for the AD banner, including the headline, subheadline, and CTA text. \\
- CTA text used in CTA button must be short catchy and concise \\
- The banner stands out when small \\
- The banner has a good background image \\

Here are some examples of prompts: \\
- Example 1 \\
- Example 2 \\
- Example 3

\end{tcolorbox}

We picked prompt examples from the AI Image Prompt Collection~\cite{imagepromptcollection} for Flux, the official prompt crafting guide~\cite{recraft_prompt_crafting} for Recraft V3, and the intention-to-design-plan dataset~\cite{opencole2024} for OpenCOLE. For DALL·E 3, as we could not find a suitable source of examples, we manually crafted them.

\end{document}